# Computational frame analysis revisited:
# On LLMs for studying news coverage

Sharaj Kunjar[1,2*], Alyssa Hasegawa Smith[1], Tyler R Mckenzie[1], Rushali Mohbe[2,3],
Samuel V Scarpino[1,2,3,4,5,6†], Brooke Foucault Welles[1,7†]

[1]*Network Science Institute, Northeastern University, Boston, MA, 02115*

[2]*Institute for Experiential AI, Northeastern University, Boston, MA, 02115*

[3]*Khoury College of Computer Sciences, Northeastern University, Boston, USA, 02115*

[4]*Department of Public Health and Health Sciences, Northeastern University, Boston, USA, 02115*

[5]*Santa Fe Institute, Santa Fe, NM, USA, 87501*

[6]*Vermont Complex Systems Institute, University of Vermont, Burlington, VT, USA, 05405*

[7]*College of Arts, Media and Design, Department of Communication Studies, Northeastern University, MA, 02115*

Computational approaches have previously shown various promises and pitfalls when it comes to the reliable identification of media frames. Generative LLMs like GPT and Claude are increasingly being used as content analytical tools, but how effective are they for frame analysis? We address this question by systematically evaluating them against their computational predecessors: bag-of-words models and encoder-only transformers; and traditional manual coding procedures. Our analysis rests on a novel gold standard dataset that we inductively and iteratively developed through the study, investigating six months of news coverage of the US Mpox epidemic of 2022. While we discover some potential applications for generative LLMs, we demonstrate that they were consistently outperformed by manual coders, and in some instances, by smaller language models. Some form of human validation was always necessary to determine appropriate model choice. Additionally, by examining how the suitability of various approaches depended on the nature of different tasks that were part of our frame analytical workflow, we provide insights as to how researchers may leverage the complementarity of these approaches to use them in tandem. We conclude by endorsing a methodologically pluralistic approach and put forth a roadmap for computational frame analysis for researchers going forward.

*corresponding author: sharaj.kunjar@gmail.com
†Equal contribution

Frame analysis is one the most prominent and popular conceptual approaches to study communicated artefacts in media research (Lecheler & De Vreese, 2018). Put briefly, it involves the study of how an issue is talked about, by examining what pertinent aspects of the issue are included[1] when communicating about it (Entman, 1993). Studying frames communicated through mass media[2] is of particular interest, as it simultaneously reflects underlying sociopolitical processes that (re)produce cultural norms and systems of power via communicative acts[3] (Vliegenthart & Van Zoonen, 2011); and harbors the potential to (re)shape how consumers of said media weigh various considerations when thinking about the issue[4] (D. Scheufele, 1999; D. A. Scheufele & Tewksbury, 2007). However, albeit its tenure and prevalence in media research, the systematic empirical identification of media frames remains a challenge (Eisele et al., 2023), with scholars diverging on both, conceptual and methodological grounds regarding what constitutes a frame and how one appropriately measures it (Cacciatore et al., 2016).

Traditionally, the content analysis of media frames has been done via manual coding of texts. This method allows for a wide range of epistemic orientations towards framing, including hermeneutical goals[5] (Matthes & Kohring, 2008), which is often a necessity when unpacking the complex narratives inherent to media framing (Nicholls & Culpepper, 2021). However, manual coding has been critiqued for being too time-consuming, costly and opaque (Törnberg, 2024b). As a response to these limitations, scholars have shifted towards using various computer assisted-approaches such as dictionary-based counting (Chinn et al., 2020), semantic networks (Calabrese et al., 2019), topic modelling (Bhatia et al., 2021; Walter & Ophir, 2019) and supervised language models (A. C. Kroon et al., 2022; Mendelsohn et al., 2024a) for frame detection. While these methods certainly appeal to ideals of replicability and scalability, their validity is often criticized for rendering overly simplistic and mechanistic outcomes (Baden et al., 2022). As a result, researchers performing content analyses of media frames are plagued with a classical tradeoff between the reliability of systematic computational methods and the validity of conventional manual coding (Jünger et al., 2022).

Recently, the advent of large language models (LLMs) has opened up opportunities for computational methods to transcend the validity-reliability tradeoff (Bail, 2024). Characterized by their distinctive abilities to digest and generate human-like speech, LLMs have demonstrated unique capabilities to automate content analytical tasks that require contextual sensitivity (Farjam et al., 2025; Törnberg, 2024b), including frame analysis (Gilardi et al., 2023; Pastorino et al., 2024). However, LLMs are not without their own limitations: heightened sensitivity to minute implementation details (Törnberg, 2024a), hallucinations that lead to spurious outputs (Farjam et al., 2025), stochasticity in their outcomes (Bender et al., 2021), difficulties in analyzing long documents (Ziems et al., 2024), to name a few. Notably, studies that have evaluated LLMs for content analysis regularly risk overusing benchmark datasets that may already be part of the model's pre-training data (Törnberg, 2024a). Thus, the effectiveness of how well these models perform *in situ*, inductive settings with the lack of benchmarks is relatively underexplored. Furthermore, a comprehensive study, that evaluates LLM-based frame analysis in comparison to both, their aforementioned computational predecessors, and manual coders, remains pending.

In this study, we revisit and contribute to the ongoing debate about computational frame analysis by evaluating several language models against a novel manually coded dataset in various parts of a frame analysis workflow. Broadly speaking, we address four research questions:

**RQ1.** How does the reliability and validity of generative LLMs compare with their computational predecessors, and traditional manual coding approaches for frame analysis?

---

[1] And what aspects are excluded
[2] i.e., media frames
[3] Dubbed as "frame building" processes within constructionist perspectives of sociology (Goffman, 1974; Van Gorp, 2007)
[4] Known as the applicability-based model of "framing effects"
[5] Where one provides a deep, context-rich interpretation of the text that goes beyond superficial observations to uncover underlying meanings

**RQ2.** Does the suitability of LLMs (compared to other techniques) depend on the nature of the frame analytical task?

**RQ3.** How does the performance of generative LLMs depend on:
   a) Different implementation strategies?
   b) The accessibility and size of the chosen models?

To that end, we tested three approaches: manual coding, discriminative language models[6] and generative large language models, across three different tasks within frame analysis: relevance classification, codebook development and frame detection. As a case study, we investigated issue-specific news frames in the coverage of the US Mpox epidemic of 2022, by analyzing 3314 news articles published over 6 months (May to October 2022) in the United States. We developed and put forth a new gold standard frames codebook and manually validated dataset classifying articles for issue-relevance and presence of frames. To conclude, we present an updated roadmap for computational news frame analysis via a comprehensive examination of existing language models and offer methodological insights about the drawbacks and opportunities for the use of generative AI in media research as a whole.

## Reconciling conceptual divisions in frame analysis

Framing is one of the most rapidly expanding theories of mass communication, drawing scholarship from a wide array of disciplines including journalism (D'angelo, 2019), sociology (Goffman, 1974), psychology (Lecheler & De Vreese, 2018), and of recently, machine learning (Eisele et al., 2023; Nicholls & Culpepper, 2021). As of September 8, 2025, Semantic Scholar returned 1,080,000 results for the term "framing", underscoring the long-standing ubiquity of this multiparadigmatic research program (D'Angelo, 2002). In his seminal and widely cited work (Entman, 1993), Entman defines framing as a communicative act that "selects some aspects of a perceived reality and make them more salient in communicating text, in such a way as to promote a particular *problem definition, causal interpretation, moral evaluation, and/or treatment recommendation*" for the issue described. Frames then, constitute four distinct elements[7] serving various functions: a determination of the scope/topic of the issue at hand; an attribution of responsibility to specific acts or actors that cause the observed issue; an evaluation of the issue with respect to cultural or personal values; and a provision of solutions to tackle it. Our study utilizes this conception of frames, and follows Matthes and Köhring's suit for frame analysis (Matthes & Köhring, 2008); which involves an assessment and description of media frames by deconstructing the four frame elements that make up the frame.

A large portion of framing research has been devoted to identifying frames communicated through news media, as they elucidate the narratives that dominate mainstream mass communication (Gamson & Modigliani, 1989). However, this task has proven to be incredibly elusive, owing to the multitude of ways in which one may conceptualize and measure frames. For starters, the distinction between equivalency and emphasis frames has sparked deliberation surrounding what should be included within the purview of frame analysis (Cacciatore et al., 2016; Vreese, 2005). Inspired from the psychological tradition of prospect theory (Tversky & Kahneman, 1981), *equivalency framing* deals with the presentation of different yet logically equivalent information to induce framing effects. For example, the messages "Quitting smoking will increase your lifespan" and "Continuing to smoke will decrease your lifespan" are logically equivalent, but have different effects on smokers (Toll et al., 2008). On the contrary, *emphasis framing* involves a selective emphasis on particular considerations and facts when communicating about an issue. Returning to our example on smoking cessation, an instance of differential emphasis could be studied between "Smoking risks your personal health" vs "Second-hand smoke risks your family's health". While scholars have argued to restrict the scope of frame analysis to equivalency framing (Cacciatore et al., 2016), strictly equivalent formulations of facts do not adequately encompass the diversity of framing tactics employed by journalists when communicating about an issue (D'angelo, 2019). Thus, emphasis framing remains integral to frame analysis, and our study focusses on the presentation of conceptually and logically distinct aspects of the issue at hand.

---

[6] Bag of words models (e.g. Naïve Bayes) and encoder only transformer models (BERT, DeBERTa)
[7] Which we'll refer to hitherto as Entman's "frame elements"

Frame analysis can be further differentiated on the basis of whether they identify *issue-specific* or *issue-generic* frames (Brüggemann & D'Angelo, 2018). Issue-specific frames are only pertinent to the issue at hand and comprise of the unique and context-specific ways in which journalists communicate. Typically, these frames are identified inductively, i.e., researchers forgo pre-determined typologies to explore frames that arise out of their interpretation of the data. To illustrate, a study on the construction of HIV/AIDS in Indian newspapers (de Souza, 2007) identified three issue-specific frames: an evaluation of the severity of the epidemic, prognostic solutions, and an identification of the high risk-group. On the other hand, issue-generic frames transcend the context at hand and can be studied in relations in various topics, over time and sometimes over different cultures (Vreese, 2005). For example, Iyengar identifies an *episodic-thematic* typology of issue-generic framing in his work on television coverage of political issues (Iyengar, 1991): *episodic frames* cover specific instance and anecdotes, whereas *thematic frames* places the issue within a broader societal context. Similarly, Benford and Snow developed an issue-generic typology of core framing tasks in social movement studies (Benford & Snow, 2000): *diagnostic frames* identify problems and their causes, *prognostic frames* articulate ideal case scenarios and solutions, and *motivational frames* declare calls for action to individuals. These issue-generic typologies have been utilized by researchers to identify frames deductively across various issues.

Each conceptualization comes with its own strengths and weaknesses: where issue-specific frames allow for an extensive level of sensitivity and details regarding an issue, this sensitivity also hampers attempts to generalize and compare over multiple issues. Scholars in computational frame analysis have advocated that researchers stick to issue-generic typologies, since the deductive nature of analysis allows for comparison between different issues and thus facilitates theory building (Eisele et al., 2023). While we agree that a context-free analysis is compelling, we also note that a restriction to adaptations of prior knowledge could be limiting (Walter & Ophir, 2019) and the relative lack of granularity within issue-generic typologies may render the analysis superficial. As is, the extent to which a frame transcends an issue is hard to exhaustively determine, since the analytical difference between specific and generic frames are "more subtle than stark" (Brüggemann & D'Angelo, 2018, p. 93). Most importantly, in our analysis of the Mpox epidemic's coverage, we observed various frames that transcended conventional issue-generic typologies, especially since a generic typology for public health communication is yet to be established. Given these caveats and our inductive goals, we opted to perform an issue-specific analysis for our case study, while acknowledging the limitations of this approach.

## Reliability and validity in frame analysis

The lack of a uniformity in the conceptual facets of frame analysis is also mirrored in its methodological debates. The quest to optimally identify media frames has been arduous, resulting in a plurality of techniques that often complement each other's strengths and weaknesses. As with any general content analytical task, these techniques have been evaluated via an assessment of their reliability and validity. Techniques are valid inasmuch as the measurement "represents the intended, and only the intended, concept" (Neuendorf, 2002, p. 112). Difficulties in the empirical determination of frames stems from ambiguous relationships between abstract meanings that underly frames and its textual contents. Researchers often rely on operational definitions through a careful reading of the texts in order to ensure the measurement of the "intended concept", i.e., validity. However, these frame definitions are contingent on individual interpretations of texts and a series of subjective judgements, making them prone to idiosyncrasies (Van Gorp, 2010). Therein lies the power of reliability: techniques are reliable insofar as they are replicable (Krippendorff, 1999, p. 18), i.e., yield the same findings on repeated trials over the same data, irrespective of the researcher, time or circumstances. By ensuring that multiple annotators yield reasonable agreement over a content analytical outcome, researchers ensure that their measurement is robust to intersubjective contestation.

Manual coding has been one of the oldest and most typical ways to identify frames. Matthes and Kohring identified three broad manual coding based approaches for inductive frame analysis (Matthes & Kohring, 2008): first, the *hermeneutic approach*, where frames are qualitatively described in depth over a relatively smaller sample size without the need for any form of quantification. Following the interpretivist tradition of social sciences (Schwartz-Shea & Yanow, 2015), researchers are encouraged to be reflexive and embrace their subjectivity to construct meaning out of the communicated text. While the requirements of reflexivity[8] and thick descriptions[9] in this approach allows for a profound level of thoroughness in the operational definition, it faces significant reliability concerns since the exact procedures to arrive at the outcome is somewhat ambiguous, and by principle, value laden. Second, the *linguistic approach*, where patterns in the occurrences of specific units in the text (e.g. words, sentences etc.) are explicitly mapped to a frame. Such a rules-based approach allows the researcher to restrict the role of individual values to the operational rules, thereby appealing to norms of procedural objectivity[10] (Douglas, 2004). However, the inordinate requirement for explicit linguistic cues makes this approach restrictive under many circumstances, e.g., polysemy[11]. Finally, we have the *manual holistic approach*, where frames are first developed qualitatively into a codebook which is then used by human annotators to categorize textual content into frames. Typically, the coders are encouraged to iterate between the text, their own interpretation and the other coders' interpretation, of the text, thus ensuring interactive objectivity[12] (Douglas, 2004). The circular nature of this approach, paired with an adherence to codebook guidelines addresses many of the issues posed by the two aforementioned methods, making it the top choice for our study.

Despite the myriads of attempts to reconcile hurdles in ensuring reliability and validity, certain issues prevail for manual content analysis of media frames. First and foremost, the scalability of this approach is limited (Grimmer & Stewart, 2013). As is often the case, frame analysis corpora span thousands of documents, and researchers only code a fraction of it. When pursued, large-scale annotation is expensive, as it requires hiring multiple human coders. Second, as evidenced by the consistent underperformance of untrained coders over domain experts (Gilardi et al., 2023), even for codebook assisted methods, domain-knowledge remains a barrier. That is, manual coders are required to develop domain expertise before analysis to ensure reliability, which in turn is laborious and time intensive. Finally, researchers run the risk of extracting "coder frames" instead of media frames (Matthes & Kohring, 2008), where their interpretation of the text remains anchored to their initial perceptions, thus compromising on validity. As a response to the aforementioned drawbacks of manual coding, scholars have been shifting to various computational approaches for frame analysis.

## A broad overview of computational frame analysis

Performing an exhaustive sweep of computational frame analysis methods is a task beyond the scope of a single paper. However, prior to the release of generative large language models, these approaches could be broadly classified into three categories (Baden et al., 2022): rule-based classifiers, unsupervised machine learning models and supervised machine learning models[13]. *Rule-based approaches* encompass a wide array of methods that categorize content based on definitive rules. For instance, in dictionary-based approaches, a text represents a certain frame if it contains terms enumerated in a pre-determined dictionary that operationally defines the frame. *Unsupervised approaches* aim to discover frames by inductively exploring regularities in the text, with very little to no pre-defined classes. Innovations in language modelling over the past decade, particularly the release of pre-trained transformer models like BERT (Devlin et al., 2019; Grootendorst, 2022), has given rise to an explosion in the use of unsupervised methods like semantic network analysis and topic modelling[14] (Walter & Ophir, 2019). Finally, *supervised approaches* use machine learning models to train classifiers on a representative gold-standard dataset to infer the implicit rules that optimally predict whether a text belongs to a certain class. This gold standard dataset, also known as the training dataset, provides the foundation and determines the performance for this approach, and is often obtained by manual coding. Albeit the requirement for intensive human input and supervision, supervised approaches have proven to be the most effective of all the preceding computational methods, and were hence prioritized in our analysis (Eisele et al., 2023).

---

[8] Reflexivity entails examining the ways in which researcher's own positionality and standpoint shapes the way they analyze data. This objective recognizes that scientific outcomes are not value-free, and that individual subjectivities are central to the research process.

[9] Introduced by Clifford Geertz in his anthropological works (Geertz, 2017), thick descriptions refer to descriptive details surrounding the communicated texts to create a nuanced portrait of the cultural layers that inform the researcher's interpretation of the text.

[10] Procedural objectivity is achieved when one could lay down systematic procedures that "impose uniformity on processes, allowing for individual interchangeability and excluding individual idiosyncrasies or judgments from processes" (Douglas, 2004)

[11] When a phrase/word has multiple co-existing meanings, e.g. "ring a bell"

[12] Also dubbed as "dialogic intersubjectivity" elsewhere (Gillespie & Cornish, 2010), interactive objectivity requires participants to go beyond agreements and coordinate their perspectives, especially when they're diverging, to develop shared interpretations of the text through dialogue and negotiation of these divergences.

[13] Note that these three approaches are by no means exhaustive or disparate. Several approaches (decision tree learning) are often a combination of some of the three approaches we've enlisted.

Digging further into supervised approaches, Kroon identifies two categories of models, demarcated by their treatment of linguistic units within texts (A. Kroon et al., 2024): *Bag-of-word (BoW) models* and *context-aware models*. BoW models ignore word order and consequently don't explicitly encode any syntactical, semantic or pragmatic structure when learning from text. For instance, Naïve Bayes classifiers (Lewis, 1998), the BoW model incorporated into our analysis, works by learning dependencies between input words and output labels (e.g. frames). Importantly, here, different words within a text are treated as independent[15], i.e., co-occurrences and correlations between words are ignored. Due to their relative simplicity, these models offer high degrees of explainability and accessibility, making them a great model choice when they work reliably. However, the assumption of independence between words undermines the complexity of human language and consequently the validity of these methods. For instance, "we are supportive of AI in education" and "we are not supportive of AI in education" are two sentences with nearly identical word distributions, but while the former represents of "pro-AI" frame, the latter doesn't. Distinguishing between those two sentences requires a model to perceive words in the context of their sentence ("supportive" in context of the preceding "not"). Such strides in modelling context-awareness was achieved via the introduction of neural networks, and in particular, the transformer architecture, that capture word meanings in relation to its surrounding text. These models are also pre-trained on vast amounts of unlabeled data, during which they acquire "general language knowledge" (A. Kroon et al., 2024), following which they require lesser training data for specific classification tasks. While these models do not offer the same level of explainability as BoW models, these models are open source, making them more accessible and transparent than the recent proprietary LLMs. Thus, these models, specifically BERT (Devlin et al., 2019) and its variants, e.g., RoBERTa (Y. Liu et al., 2019), DeBERTa (He et al., 2021) , S-BERT (Reimers & Gurevych, 2019), have been widely adopted within computational frame analysis communities (Eisele et al., 2023; Jumle et al., 2025; Mendelsohn et al., 2024a). Note that unlike the latest generative large language models like GPT, the aforementioned ones are encoder-only[16] transformer models, i.e., they encode input texts but do not generate new output texts. Due to this caveat, in the rest of the study, we refer to encoder-only transformer models and BoW models as *discriminative language models[17]*.

---

[14] Also, combinations with prior methods, e.g., keyword-assisted topic models, analysis of topic model networks (ANTMN)

[15] Given word counts x and y for words X and Y, the probability that an article A contains a frame F:
$P(F \in A | n(X) = x, n(Y) = y) = P(F \in A | n(X) = x) \times P(F \in A | n(Y) = y)$.

[16] Encoder-only models are designed to cipher texts in an embedding space by processing chunks of text (called tokens) in relation to each other and build an understanding of the input text based on its location in the embedding space. These models are useful for a wide array of text classification tasks, e.g., sentiment analysis, since they can arguably situate meaning of a text by encoding its tokens in an embedding space. These models are in contrast to decoder-only models, such as ChatGPT, which generates new text based on the input. Unlike encoder-only models, decoder-only models are unidirectional, in that tokens are processed sequentially, with future tokens being determined by preceding tokens, but the not the other way round.

[17] Two caveats: First, BoW models are not language models in the conventional sense since, unlike neural networks, they don't encode language via word/token embeddings or perform next-token prediction. Second, encoder-only transformer models such as BERT can be turned into a generative LLM (Samuel, 2024) with extra steps, e.g., adding a generation head. We adopt "discriminative language models" only for ease of use and to differentiate them from the off the shelf "generative pre-trained transformer models" such as GPT, Claude, Gemini etc.

In spite of the many revolutions in computational methods, some issues in reliability and validity persist. Even with heightened context-awareness, the aforementioned approaches fall short of manual coding in tasks which require pragmatic understanding, like detecting sarcasm, sentiments and stances (van Atteveldt et al., 2021), signifying a broader tradeoff between automation and interpretation (Jünger et al., 2022). Human coder biases within conventional techniques is retained in supervised approaches (Laurer et al., 2025), since these models are trained on researcher-generated data before deployment. Furthermore, unlike conventional dictionary-based techniques, implementing supervised models require a fair amount of computational literacy and access to good-quality training data, and hence poses some accessibility concerns. Thus, while computational methods made significant advances over manual coding in terms of scalability and reliability, limitations in their validity and accessibility lingers.

## Could generative AI transform computational frame analysis?

The recent emergence of generative AI, particularly marked by the release of ChatGPT in 2022, has offered hopes in overcoming many of the aforementioned limitations in computational approaches. Architecturally, these models are *decoder-only* (Radford et al., 2018), meaning that they predict (decode) the next token in a text based on the input, thus leading to their colloquial name – generative large language models. Many qualities of generative LLMs stand out as novel and promising: first, they are instruction-tuned, meaning that they can follow conversational instructions provided by the user to perform tasks and generate natural language outputs (Törnberg, 2024b). Second, they are capable of *zero-shot* learning, which allows them to function solely based on the provided set of instructions and no further training, circumventing the need for massive amounts of training data and consequently the biases that arise out of imbalances in the user-generated training data (Laurer et al., 2025). Third, along with text generation, these models have extraordinary reasoning abilities, allowing them to simultaneously provide chain of thought justifications for their classification decisions and retrospect over these decisions post-hoc. This ability has been leveraged by scholars to design agentic workflows (Farjam et al., 2025), where LLMs could improve upon the validity of their outcomes by evaluating their chain of thought. As a result of many such innovations, generative LLMs have proven to be effective in various text-annotation tasks, often surpassing (untrained) crowd-workers (Gilardi et al., 2023) and sometimes, expert human coders (Törnberg, 2024b).

Although the application of generative LLMs for frame analysis is relatively underexplored, some scholarly work has highlighted potential integration of these tools in various parts of the pipeline. At the outset, frame analysis often involves relatively simplistic but laborious processing tasks associated with curating a dataset, such as classifying documents for relevance based on particular inclusion criteria. While relevance classification can be automated to a certain extent with discriminative language models (Mendelsohn et al., 2024b), Gilardi and Kwon demonstrate a more economical and efficient alternative for the same with generative LLMs (Gilardi et al., 2023; Kwon et al., 2025). Next, inductive frame analysis workflows entail a codebook development stage, where researchers identify issue-specific frames within the corpus. Where conventional computational approaches (e.g. topic modelling) have had validity concerns in this regard, recent work on performing LLM-in-the-loop codebook development illustrates a prospective use of generative LLMs to identify frames with high interpretive capacity (Dai et al., 2023; De Paoli, 2023).

Finally, a number of scholars have examined the potential of using generative LLMs to perform content analysis of media frames. In a frame detection test on the gun violence frames corpus[18] (S. Liu et al., 2019), Pastorino et al observed moderate performances[19] using GPT-3.5 and GPT-4 in zero-shot and few-shot settings, with models struggling when headlines become nuanced and emotional. They also found that smaller fine-tuned models, e.g., FLAN-T5, often performed better than the GPT models. Most recently, Kwon et al showed that while GPT-4 performed satisfactorily in detecting frames related to COVID-19 in press releases, prompt engineering and document length were major factors in its effectiveness (Kwon et al., 2025). In contrast, Alonso del Barrio et al found that GPT-

---

[18] A benchmark dataset of issue-specific frames in headlines from US newspapers about gun violence
[19] Highest F1 scores between 60-70%

3.5 performed poorly[20] when detecting issue-generic frames in television transcripts (Alonso Del Barrio et al., 2024). Adding to the debate, while Gilardi et al found that GPT-3.5 consistently outperformed crowd workers in all frame detection tasks, it exhibited highly variable performance[21] depending on the dataset. As such, these studies show a somewhat mixed bag of results based on the nature of the task, with generative LLMs performing well on more mechanistic tasks, but worse on nuanced classification tasks. To add to the confusion, LLM performances clearly depended on implementation and model selection details in ways that weren't predictable.

Despite the brimming prospects, many methodological challenges regarding generative LLMs warrant careful consideration. Unlike the discriminative models we evaluated, not all LLMs are free for use. In general, proprietary models tend to outperform the open-source ones (Huang et al., 2024), underscoring concerns about accessibility. As such, the pre-training procedures for generative LLMs are also not transparent, thus making them black boxes (Ziems et al., 2024) with very little explainability. Relatedly, the outputs of these models vary significantly based on minute implementation details (e.g., prompts, hyperparameter selection), making it unacceptably simple to hack them to obtain the output a researcher prefers (Baumann et al., 2025). In addition, LLM outputs tend to be stochastic and prone to hallucinations[22] (Bender et al., 2021; Tan et al., 2024), posing significant risks to reproducibility. These concerns also tend to stack up as the input texts get longer (Ziems et al., 2024), raising questions about their effectiveness for studying news articles, which are the object of study for news frame analysis. Additionally, the capabilities of generative LLMs to perform highly interpretive tasks remains a contested topic (Choi et al., 2023; Fan et al., 2024; Sravanthi et al., 2024), and to say the least, when the task is complex, they run into similar issues of biases and skewed interpretations as traditional manual annotation procedures (Brown et al., 2025). If anything, in a pursuit to achieve value-freedom and a 'view from nowhere', LLMs often inferiorize non-Western and marginalized standpoints (Mollema, 2025). Recent work (Samuel, 2024) has also found instances where simpler and older transformer models outperform generative LLMs. As such, these findings warrant a more comprehensive and comparative investigation into the performance of these models with respect to their predecessors and traditional techniques, thus underscoring the relevance of our study. We hope to address some of the transparency concerns by laying bare the prompts and implementation choices made in the study. Our study design also allows for a human validation of the model outputs, thus avoiding a deferral of value-laden analysis onto LLMs.

# Data

Our study examines issue-specific news frames in the coverage of the Mpox epidemic in the United States as a case study. Marked by a spillover of the Mpox[23] clade II virus from endemic regions in central and western Africa in 2022, this epidemic disproportionately affected Black and Hispanic communities in the US (Philpott, 2022). Moreover, although Mpox is not classified as sexually transmitted infection (STI), this clade of the virus overwhelmingly concentrated in sexual minorities, with over 95% of the cases occurring among Men who have Sex with Men (MSMs) (Kupferschmidt, 2022). Given its intersection with controversies surrounding race, sexuality, public health and global relations, the media discourse about the epidemic was heavily politicized (März et al., 2022). Studies on social media (Anoop & Sreelakshmi, 2023; Hong, 2023; Owens & Hubach, 2023) revealed a prevalence of fear and stigma-based messaging, anti-LGBTQIA+ rhetoric and distrust in public health institutions. Given the role journalistic framing has played in constructing and legitimizing mainstream perceptions surrounding other public health topics as COVID-19 (Ogbodo et al., 2020; Teschendorf, 2024) and HIV/AIDS (D'Angelo et al., 2013; de Souza, 2007), a frame analysis of the news coverage of the Mpox epidemic seemed promising in terms of the variety of salient narratives we could discover. While we reserve the discussion of particularities about this discourse for our forthcoming work, some of these narratives are summarized in our frame detection codebook (see Appendix section A3.2.1). The potential to discover a variety of contentious media frames, in tandem with the relative accessibility of US news articles surrounding the Mpox epidemic, made it the top choice for our case study.

---

[20] Accuracy of 43%
[21] Accuracies between 30% to 80% depending on the dataset
[22] Outputting very confident but incorrect information that is absent in the pre-training dataset
[23] Formerly known as "monkeypox"

Our final dataset consisted of 3314 English language news articles about the Mpox epidemic, published in a 6-month period between May 1 and October 31, 2022. This six-month period comprised of some of the most noteworthy developments of the epidemic – ranging from the first known US case of the virus on May 18[th], the WHO declaration of a public health emergency on July 23[rd], and the peak of the epidemic over August 2022. These articles were published as online news by 131 unique mainstream news outlets that cover media largely about the United States. All the articles were written in English. We obtained this corpus by querying for articles that contain the terms "mpox" or "monkeypox", from Media cloud (Roberts et al., 2021), an open source database of online news; and further web-scraping texts and other metadata about the articles using the python package newspaper4k (Paraschiv, 2024). Before curating the final dataset, the articles were subject to various pre-processing and cleaning steps: First, articles that didn't meet the inclusion criteria (e.g., word counts) were removed via simple string evaluation. Second, to maximize variation in the texts, we deduplicated the dataset by removing copies of almost identical articles. Third, we cleaned up the texts for each article by removing non-editorial content, such as advertisements and letters to the editors. The details for each of these steps can be found in the Appendix section A1.2.

## Methods

To address **RQ1**, our study compared the effectiveness of three different approaches in frame analysis: manual coding, discriminative language models and generative large language model, all of which will be elucidated in the following sections. The study design allowed us to test these approaches in various parts of the workflow: First, **relevance classification**, where we processed our dataset prior to frame analysis to include only those articles that substantially talk about the Mpox epidemic. While pre-processing texts based on the presence of certain keywords helped narrow down the dataset, we wanted to exclude articles where Mpox was not the primary focus, e.g., an article about gay marriage that references stigma-based messaging during Mpox as an example of homophobia. Second, **frames codebook development**, where we developed a set of Mpox-specific frames with definitions and examples. Unlike issue-generic frame analysis, where one could deductively code frames based on a pre-determined typology, our analysis aimed to uncover frames that were specific to our context, thus allowing us to test the potential of the aforementioned approaches in identifying new frames from scratch. Finally, **frame detection**, where for each article, we labelled if each of the frames, as defined by the codebook, occurred in the texts.

These three tasks require very distinctive capabilities: relevance classification is a relatively straightforward content analytical task where simpler measures such as word counts (e.g., "mpox" being mentioned more than 3 times) could serve as a good proxy for relevance. Frame detection, on the other hand, is a content analytical task requiring varying depths of interpretation depending on the complexity of the frame, thus demanding a more hermeneutical approach towards understanding language. As such, the codebook for frame detection was also more extensive and detailed than that for relevance classification, demanding a more careful categorization effort. Finally, frames codebook development was primarily a summarization task. As opposed to the content analytical tasks where one could refer to a codebook for common ground, codebook development was entirely inductive and arguably the most subjective task, since there are a myriad of ways of thematically identifying and organizing ideas from a corpus. Thus, in addition to evaluating how LLMs could augment different chronological stages of the analysis, the three-tier design allowed us to address **RQ2** by assessing how different frame analytical approaches perform based on the nature of the task.

We evaluate our approaches based on a few metrics commonly used in computational communication research (Burscher et al., 2014; Eisele et al., 2023): Cohen's Kappa ($\kappa$), percent agreement, F1, precision and recall scores. Percent agreement (also known as accuracy) denotes the raw proportion of the data where the annotator and the gold standard agree upon on their annotations. Cohen's Kappa contextualizes this percent agreement in terms of what the agreement would have been by random chance. We interpret the $\kappa$ scores based on Landis and Koch's criteria[24] for the content analysis of categorical data, where generally, an approach that leads to a Kappa score above 0.6 is considered reasonably reliable (Landis & Koch, 1977). Precision is the proportion of positive predictions that were also labelled as positive by the gold standard technique, thus indicating the extent to which the approach renders false positives. In contrast, recall is the proportion of negative predictions that were labelled as negative by the gold standard, indicating the extent to which the approach renders false negatives. Finally, F1 score is the harmonic mean[25] of precision and recall, with F1 scores above 0.7 generally being an indicator for good performance.

## Manual coding

All manual coding was performed by the first and the third authors of the study. They performed relevance classification by utilizing a codebook-based iterative annotation procedure. First, they independently combed through the articles at a high-level to identify inclusion criteria for relevance. Second, they discussed their independent observations and developed a codebook that operationalized relevance (see Appendix section A2.1.1 for relevance codebook). Finally, they sampled a random subset of 500 articles[26] for relevance classification and proceeded to manually label each article as relevant/irrelevant based on the developed relevance codebook. After two rounds of annotation, they achieved an inter-rater percent agreement 98.8% and a Cohen's $\kappa = 0.97$. The first author's annotations after the second round of coding were then subsequently used as the gold standard labels.

After obtaining relevance labels for the entire dataset, the manual coders adopted an **applied thematic analysis** (Guest et al., 2012; Mackieson et al., 2019) procedure on a random subset of a 100 articles[27] to develop a codebook for frame detection (see Appendix section A2.2.1 for details). The affordance to accommodate a diverse range of epistemological orientations, along with its previously documented utility in identifying narrative frames in public health research (Tahamtan et al., 2021), made thematic analysis the top choice among manual coding techniques for our study. At this stage, we chose an *inductive*, *interpretive* and *iterative* approach. That is, the coders were encouraged to develop insights based on patterns in the data (inductive), look beyond surface and lexicon-level interpretations of the text (interpretive) and achieve intersubjectivity by requiring them to develop a shared understanding of the frames over multiple rounds of close reading (iterative). Furthermore, to improve reliability and consistency in frame definitions, they followed Matthes and Kohring's procedure (Matthes & Kohring, 2008) of elucidating Entman's elements (Entman, 1993) for each frame – problem definition, causal interpretation, moral evaluation and treatment recommendation. This procedure resulted in a codebook with seven frames (see Appendix section A3.2.1 for frames codebook).

Finally, the manual coders annotated a subset of 100 articles for the presence of the seven frames elucidated in the codebook developed above. The procedure was one-to-many and at an article level, i.e., for each article, the annotators labelled whether each frame was present at least once in the entire article. At this stage, to improve reliability, the annotators were encouraged to look for explicit presence of frames based on their codebook definitions and avoid using inferred connotations as a basis for the presence of different frames. After two rounds of manual coding, we achieved an inter-rater percent agreement of 95.14% and Cohen's $\kappa = 0.84$. Subsequently, these 100 articles were used as the test set for language models, with the first author's labels as gold standard. Furthermore, in order to train the aforementioned language models, the first author labelled an additional random subset of 400 articles.

---

[24] $0 < \kappa \leq 0.2$ (slight), $0.2 < \kappa \leq 0.4$ (fair), $0.4 < \kappa \leq 0.6$ (moderate), $0.6 < \kappa \leq 0.8$ (substantial), $0.8 < \kappa < 1$ (near perfect)

[25] $F1 = \frac{2 * Precision * Recall}{Precision + Recall}$

[26] Out of 3314 articles in the dataset post pre-preprocessing

[27] Out of 2224 articles that we obtained as relevant to the Mpox epidemic post relevance classification

## Discriminative language models

Using the gold standard labels acquired from manual coding, we trained and evaluated three discriminative language models: Naïve Bayes (Lewis, 1998), BERT-base-cased (Devlin et al., 2019), and DeBERTa-v3-base (He et al., 2021). These models were chosen based on their common use in computational communication research. Naïve Bayes, a bag-of-words classifier, was used as a baseline to test how well models could classify articles based on lexical cues. BERT, one of the most widely used encoder-only transformer models, was selected due to its previously documented performances in frame analysis (S. Liu et al., 2019). DeBERTa, one of the popular BERT variants, was chosen due to its demonstrated improvements over BERT (He et al., 2021). The specificities surrounding how each model was trained and deployed can be found in Appendix section A2.1.2.

First, we tested evaluated models on the relevance classification task against the 500 articles that were manually labelled for relevance. The Naïve Bayes classifier was trained using a 300/200 train/test split and the transformer models were trained with a 200/150/150 train/test/evaluation split. Second, following prior work on unsupervised topic modelling for discovering issue-specific frames (Bhatia et al., 2021), we used BERTopic (Grootendorst, 2022) on the entire relevant set (2224 articles) to discover high level topic representations. Finally, we performed frame detection against the 100 articles that were coded by both annotators. Here, the Naïve Bayes classifier was trained on 400 manually coded articles, and transformer models were trained on a 240/160 train/evaluation split on the same dataset. While we acknowledge that these training sets are relatively small, previous research has documented that training sets of a few hundred examples are sufficient to automate framing analysis with supervised approaches (Burscher et al., 2014).

## Generative large language models

To test generative LLMs, we primarily worked with Llama 3.1-8B-Instruct with 8-bit quantization[28] (Grattafiori et al., 2024), executed on a consumer grade local machine (details in Appendix section A2.1.3). When we conducted the study, the Llama models generally outperformed several other state-of-the-art open source models (Farjam et al., 2025). Their ease of deployment via Ollama (*Ollama*, 2023), an open source LLM application, also makes them more accessible for non-intensive programmers. Unlike most proprietary LLMs such as ChatGPT, we were able to use the Llama models unencumbered by monetary and API rate limits. Moreover, in contrast to running models on cloud servers, running them locally allowed us to maintain privacy over our data and manual labels. Thus, the Llama models were our primary choice for experimentation with most tasks.

Answering **RQ3a** required us to explore a variety of implementation strategies. However, given the plethora of options surrounding the same, an exhaustive test of the same was beyond the scope of a single study. We primarily report performances on fine-tuned and zero-shot settings with Llama and defer the reader to Appendix sections A2.3.4 and A3.3.4 for other strategies we tested. After prompt-engineering and adhering to recommended guidelines (Schulhoff et al., 2025; Törnberg, 2024a), we settled on a set of prompts to instruct our LLMs for downstream tasks (e.g., see **Figure 1**).To address **RQ3b**, we took evaluated two models: first, we tested with GPT-OSS-20B[29] (OpenAI et al., 2025), the latest open-source models from OpenAI at the time of the study. While these models are free to use, they're too large to run on a typical consumer grade laptop with Ollama and hence exhibit a moderate level of accessibility. Second, in spite of their proprietary and inaccessible nature, we managed to test Claude Sonnet 4.0[30] (Anthropic, 2025), since we did not want to entirely discount the capabilities of LLMs based on feasibility concerns. However, due to their heftier computing requirements and monetary restrictions, we only tested Claude and GPT-OSS in the zero-shot setting. Thus, we address **RQ3** by evaluating three models with differing levels of accessibility across two implementation strategies.

---

[28] Henceforth referred to as Llama
[29] Henceforth referred to as GPT-OSS
[30] Henceforth referred to as Claude

First, we evaluated how LLMs performed on the relevance classification task. In the zero-shot setting, the models were tested against all the 500 articles that were manually labelled for relevance. For the fine-tuned setting, we trained (and tested) on the same 300/200 train/test split as the Naïve Bayes model. Second, we evaluated their capabilities in generating a new frames codebook based by utilizing an LLM-based inductive thematic analysis framework (Dai et al., 2023; De Paoli, 2023). Similar to the manual codebook development procedure, we utilized a subset of 100 relevant articles to sequentially generate initial codes, cluster these codes into themes, and then review the themes to define frames (for details, see Appendix section A2.2.3). Finally, we evaluated LLM performance on the frame detection task. For both the zero-shot and fine-tuned settings, the models were evaluated against a test set of 100 articles that were manually coded by both annotators. For the fine-tuned setting, we trained the models on 400 articles.

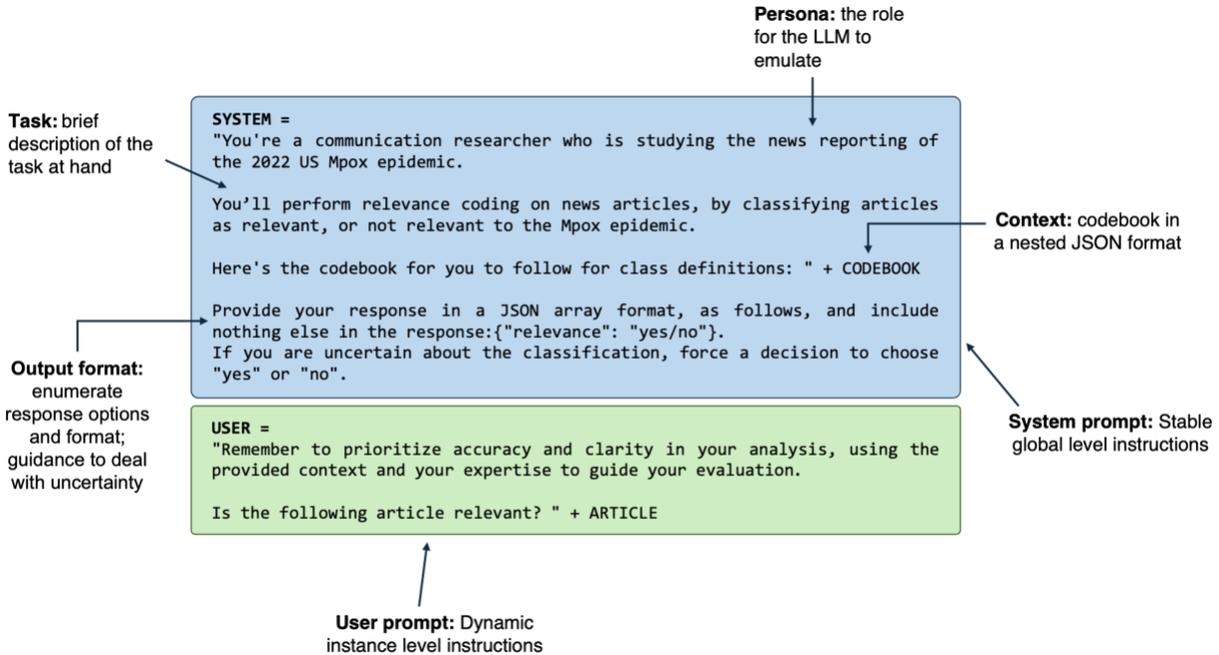

**Figure 1:** Sample structured LLM prompt, split into system and user level instructions.

# Results

Using the first author's manual annotations as the gold standard, we evaluated methods for three types of tasks – relevance classification, frames codebook development, and frame code detection, using three broad approaches: manual (human) coding, discriminative language models and generative large language models. Among all the language models, the Naïve Bayes model acted as the baseline, with the least amount of sophistication in its treatment of language, but high levels of explainability. The approaches that followed: BERT, DeBERTa, Llama, GPT-OSS and Claude all represent language models of increasing size and thus, decreasing accessibility, in that order. Generally, we found that manual coding outperformed all the other approaches across all the tasks. For simpler tasks, discriminative models offered a more accessible, and often more reliable option than their generative successors. For more complex tasks, generative LLMs outperformed their predecessors, but only achieved moderate agreement scores with the gold standard, thus highlighting the need for careful human validation when using them. Moreover, proprietary models consistently outperformed the open-source ones, raising further concerns about the tradeoff between accessibility and reliability.

## Relevance classification

Human coders manually classified articles for relevance over two rounds of iterative coding, yielding a Cohen's $\kappa = 0.82$ in the first round, and $\kappa = 0.97$ in the second. They found that 64.2% of all pre-processed articles were relevant to the Mpox epidemic. While they consistently outperformed all the language models we evaluated (**Table 1**), all the models achieved near perfect agreement ($\kappa \geq 0.8$) with the gold standard, with DeBERTa emerging as the most reliable model. An assessment of the precision-recall scores (Appendix section A3.1) reveals no significant imbalance between false positives and negatives within all model predictions. Notably, every discriminative language model, including our Naïve Bayes classifier, achieved near perfect agreement with the gold standard and outperformed generative LLMs. This finding is contextualized by the caveat that the category (relevance) was reasonably tractable by patterns in lexical units such as words and phrases. For example, an article that contained the word "monkeypox" five times is easily much more likely to be relevant than an article that mentions it once. The superior performance of the Naïve Bayes classifier highlights the power of simpler, explainable models over more complex, generative ones for straightforward tasks, potentially also offering solutions to ease the cognitive load on manual coders. While the generative LLMs generally underperformed their discriminative counterparts, we note that the zero-shot performances of the models only marginally trailed behind the fine-tuned ones. Furthermore, the open-source models also achieved competitive performances. Thus, zero-shot classification with open-source generative LLMs could be a viable option for relevance classification when one is unable to manually code a training dataset or reliability is lesser of a priority.

| Annotator/Metric | Percent Agreement (%) | Cohen's $\kappa$ | F1 score |
|---|---|---|---|
| Humans (round 1) | 92 | 0.82 | 0.94 |
| Humans (round 2) | 99 | 0.97 | 0.99 |
| Naïve Bayes | 96 | 0.90 | 0.97 |
| BERT | 96 | 0.91 | 0.97 |
| DeBERTa | **98** | **0.95** | **0.98** |
| Llama (zero-shot) | 92 | 0.81 | 0.94 |
| Llama (fine-tuned) | 95 | 0.89 | 0.95 |
| GPT (zero-shot) | 92 | 0.80 | 0.94 |
| Claude (zero-shot) | 95 | 0.88 | 0.96 |

**Table 1:** Relevance classification metrics for all annotators. The metrics for the best-performing language model is highlighted. For manual coding, the metrics are calculated by evaluating annotations between the two coders in each round. All language model predictions are evaluated against the labels produced by the first author after round 2 of manual annotation. Precision and recall scores can be found in Appendix Table A1 in section A3.1.

## Frames codebook development

Manual applied thematic analysis yielded a codebook with 7 frames (**Table 2**) of varying amounts of prevalence within the dataset. Frames A and E were the most prominent frames, owing to the debates surrounding the disease concentration in MSM networks, and diverging attitudes towards the need for public health interventions, respectively. Frames D and G also cropped up often, contextualized by the COVID-19 pandemic fatigue and general dissatisfaction with healthcare systems in 2022. Finally, while the frames B, C and F occurred infrequently in the dataset, they conveyed other conceptually distinct considerations such as the racial and global inequities in public health measures, and individual anecdotes humanizing the at-risk groups.

Based on our interpretations of the topic representations output by BERTopic (Appendix section A3.2.2, Table A2), this method generated topics that overlapped with 4 out of 7 of the gold standard frames. However, consistent with prior work highlighting validity concerns surrounding topic modelling-based approaches for the inductive identification of frames (Ali & Hassan, 2022), the discriminative models only describe frames as broad topics, without coherent frame elements. Moreover, since the topic representations comprised only of a list of keywords, their scope of use was restricted to the provision of seeds based on clustering articles[31] that require further interpretation and elucidation to develop a codebook. Thus, at least on their own, they make an unsuitable replacement for the inductive identification of frames by manual coders.

Similarly, the LLM-based inductive thematic analysis yielded a codebook of 11 frames (see Appendix section A3.2.3), which aligned well with 5 out of 7 of the gold standard frames. But, in contrast to their discriminative predecessors, they provided free-form descriptions and natural language labels for each frame, rendering their outputs significantly more interpretable. This can be attributed to their well-documented summarizing and text-generation capabilities (Ziems et al., 2024). However, as noted by prior work on using GPT to inductively identify content analytical codes (Carius & Teixeira, 2024), the hermeneutical aspects of their outcomes remain questionable, and as such, semantically similar phrases may still stem from diverging interpretations. For example, the LLM-inferred frame "Global health cooperation" may be consistent with the human-identified frame "Global relations", but while the former diagnoses challenges to global healthcare access, the latter prescribes a broader call for action encompassing both isolationist and globalist views about the same; and may thus lead to different outcomes when utilized for content analysis. Indeed, when the same model is used for downstream frame detection with two semantically similar codebooks (see Table A3, Appendix section A3.2.3), one generated by the model and the other by humans, the labels diverged. This discrepancy indicates that the codebooks produced by humans and generative LLMs are not interchangeable, even when they seem similar on the surface. Thus, although generative LLMs offer a considerable improvement in codebook development over their discriminative counterparts, we maintain the recommendations about computational approaches made elsewhere (Farjam et al., 2025; Walter & Ophir, 2019): the final hermeneutic interpretation of the outcomes remains the work of expert manual coders.

| ID | Frame | Description | Prevalence |
|----|-------|-------------|------------|
| A | Sexual stigma and transmission routes | Addresses the role of sexual identity and sexual transmission in shaping Mpox spread, and evaluates health communication strategies in that regard | 38% |
| B | Racial disparities and stigmatizing name | Addresses racial health inequities in Mpox cases/treatment and debates whether the name "monkeypox" is racially stigmatizing | 12% |
| C | Global relations | Discusses international strategy in epidemic management, ranging from international co-operation (globalism) to prioritizing national interests (isolationism) | 10% |
| D | Public health failure | Reports inadequate public health measures, highlighting bureaucratic obstacles, insufficient resources, or institutional incompetence in handling the Mpox outbreak | 19% |
| E | Epidemic preparedness and surveillance | Debates appropriate response levels and institutional trust, contrasting pro-establishment calls for aggressive action versus anti-establishment resistance to interventions. | 31% |

---

[31] similar to initial codes in applied thematic analysis

| F | Human-interest stories | Features personal narratives and stories from affected communities, emphasizing human experiences rather than statistical or structural analysis. | 9% |
| G | Broader health issues | Places Mpox within larger public health contexts, discussing its relation to broader concerns such as the COVID-19 pandemic, the AIDS epidemic, climate change etc. | 21% |

**Table 2:** Gold standard frames developed by manual coders through applied thematic analysis. Prevalence is calculated over the 100 articles that were labelled by manual coders during the frame detection step. For details descriptions with Entman's frame elements, refer to codebook in Appendix section A3.2.1.

## Frame detection

Overall, we found that manual coding provided the most reliable results (**Table 3**), highlighting its indispensability to media framing research. They achieved nearly substantial ($\kappa \sim 0.6$) agreement in their first round of annotation, surpassing most language models, and only marginally underperforming Claude. In their second round, they improved their performance to achieve near perfect agreement ($\kappa \geq 0.8$), signaling that intersubjective discussion and co-ordination of diverging perspectives between iterations of manual coding was vital to the method's superiority over other approaches. Intersubjectivity is a hallmark of content analysis (Krippendorff, 1999), and we echo the importance of dialogical intersubjectivity (Gillespie & Cornish, 2010) in developing a shared understanding of abstract concepts in social sciences, such as frames. While the manual coders exhibited slight variation in reliability scores for different frames, overall, they yielded competitive performances for every frame by the second round of annotation (**Table 4**). However, poorer performances on some frames (C, E, G) during the first round of coding reveals how this approach could also be used to determine the straightforwardness of the task; as such, if iterative coding and intersubjective dialogue is necessary to achieve substantial outcomes on a frame, then the framing task can be classified as complex.

Moving onto the discriminative models, Naïve Bayes, our baseline classifier, generally fell short for detecting frames. Its best performance was on frame A (sexual stigma and transmission routes), where it produced moderate agreement with the gold standard labels and outperformed the generative LLMs. We attribute this observation to the fact that frame A was the most prominent class in our training data while simultaneously being lexically determinable enough to yield substantial performance in our first round of manual coding. Thus, BoW models for frame analysis may be appropriate when the following conditions are met: computational resources are limited; streamlined, scalable and explainable models are preferred; and the frames can be adequately defined via distributions of lexical units. As expected owing to their context-awareness, our two encoder-only transformers surpassed the BoW model. But they marginally trailed behind the open-source generative LLMs and severely lagged behind the proprietary model. Interestingly however, this lag was not uniform across all frames. DeBERTa was the most reliable model for frames A (sexual stigma and transmission routes) and E (epidemic preparedness and surveillance), both classes that were overrepresented in our training data[32]. Attempts to mitigate class imbalance via focal loss (see Appendix section 2.3.4 and Appendix section 3.3.4) led to some improvements for the Naïve Bayes classifier, but deteriorated performance for the encoder only models. Thus, based on our findings, encoder-only transformers present a feasible, scalable and often state-of-the-art tool for frame analysis, especially for the classification of high frequency categories.

Finally, the generative LLMs achieved the highest agreement with the gold standard for frame detection among all our language models, with Claude exhibiting substantial scores ($\kappa > 0.6$), and the open-source ones exhibiting moderate scores ($0.45 < \kappa < 0.52$) on aggregate. These findings do indicate a potential for the use generative AI in frame detection. But, beyond their aggregate superiority over other models, generative LLMs warrant several methodological considerations. First, the zero-shot models only marginally lagged behind fine-tuned models, casting doubts on the effectiveness of fine-tuning LLMs, and at the least, delineating the benefits of the fine-tuning paradigm from the uncontested utility of training for supervised discriminative models. Second, when manual coders struggled to reliably identify frames, as indicated by poorer performance in the round 1 of manual coding on certain complex frames[33], the generative LLMs were also ineffective. As discussed earlier, intersubjective dialogue in manual coding was key to developing a shared understanding of the frames and attempts to emulate this procedure with fine-tuning or providing human feedback (refer to "Human feedback" in Appendix sections A2.3.4. and Table A14, A3.3.4) failed to achieve the same outcomes. As a result, even Claude, the most reliable model, hit a performance ceiling that was comparable to that of manual coders in their first round of coding. Third, beyond this barrier in coordinating diverging interpretations, the open-source generative LLMs disproportionately produced type 1 errors, leading to a large number of false positives. Finally, while Claude consistently outperformed its open-source counterparts, it produced near perfect agreement with the gold standard only on some frames (e.g. Racial disparities and stigmatizing name), where the performances of open-source models were decent to begin with. Thus, the utility of proprietary models over the open-source ones is limited to improvement over already discernable classes, as opposed to innovative performances over classes that are otherwise indiscernible with open-source approaches.

---

[32] With a prevalence higher than 30%

| Annotator/Metric | Percent Agreement (%) | Cohen's $\kappa$ | F1 score |
|---|---|---|---|
| Humans (round 1) | 86 | 0.57 | 0.66 |
| Humans (round 2) | 95 | 0.84 | 0.87 |
| Naïve Bayes | 83 | 0.28 | 0.43 |
| BERT | 85 | 0.41 | 0.58 |
| DeBERTa | 86 | 0.45 | 0.53 |
| Llama (zero-shot) | 81 | 0.46 | 0.59 |
| Llama (fine-tuned) | 82 | 0.48 | 0.61 |
| GPT (zero-shot) | 80 | 0.52 | 0.65 |
| Claude (zero-shot) | **87** | **0.63** | **0.72** |

**Table 3:** Frame detection metrics for all annotators, averaged over all frames. The metrics for the best-performing language model is highlighted. For manual coding, the metrics are calculated by evaluating annotations between the two coders in each round. All language model predictions are evaluated against the labels produced by the first author after round 2 of manual annotation. For more details see Appendix section A3.3.

| Frame | A | B | C | D | E | F | G |
|---|---|---|---|---|---|---|---|
| Prevalence (%) | 38 | 12 | 10 | 19 | 31 | 9 | 21 |
| Humans (round 1) | 0.66 | 0.74 | 0.39 | 0.64 | 0.48 | 0.64 | 0.46 |
| Humans (round 2) | 0.87 | 0.90 | 0.71 | 0.86 | 0.74 | 0.94 | 0.84 |
| Naïve Bayes | 0.55 | 0.23 | -0.02 | 0.4 | 0.42 | 0.24 | 0.11 |
| BERT | 0.57 | 0.77 | 0 | 0.34 | 0.41 | 0.43 | 0.33 |
| DeBERTa | **0.61** | 0.45 | 0.24 | 0.6 | **0.52** | 0.48 | 0.22 |
| Llama (zero-shot) | 0.43 | 0.75 | 0.24 | 0.64 | 0.36 | 0.35 | 0.45 |
| Llama (fine-tuned) | 0.42 | 0.72 | 0.24 | 0.64 | 0.47 | 0.43 | 0.45 |

---

[33] C (Global relations), E (Epidemic preparedness and surveillance) and G (Broader health issues)

| | | | | | | | |
|---|---|---|---|---|---|---|---|
| GPT (zero-shot) | 0.51 | 0.85 | 0.25 | 0.66 | 0.33 | 0.75 | 0.31 |
| Claude (zero-shot) | 0.47 | **0.95** | **0.51** | **0.68** | 0.42 | **0.88** | **0.51** |

**Table 4:** Cohen's $\kappa$ score for all annotators for different frames. The scores for the best-performing language model is highlighted. For manual coding, the scores are calculated by evaluating annotations between the two coders in each round. All language model predictions are evaluated against the labels produced by the first author after round 2 of manual annotation. For more details see Appendix section A3.3.

## Summarizing the findings

Our research questions interrogate the relationships between various approaches to frame analysis tasks. **RQ1** asked how LLMs compare with their computational predecessors and traditional manual coding, and **RQ2** asked if this comparison changes with the task type. These two questions are best answered in tandem since the evidence suggests that the suitability of different approaches depends on the nature of the task. Only one finding was an exception; akin to their discriminative predecessors, generative LLMs trailed behind manual coders across all tasks (**Table 1, 3**). Otherwise, the performance of generative LLMs were satisfactory on simpler text classification tasks (e.g. relevance) but inadequate for more complex interpretive tasks (frames A, C, E, G in **Table 4** and Appendix section A3.2.3). Notably, barring high frequency classes, generative LLMs offered significant improvements over older simpler models (**Table 3**). With regards to implementation strategies (**RQ3a**), we found that zero-shot models only marginally trailed their fine-tuned counterparts (**Table 3, 4**). Similarly, the improvements in performances obtained via error mitigation strategies (e.g. using an LLM or human as a judge, decomposing the task into indicator-based questions in Appendix section A3.3.4) was too marginal to warrant the additional effort that may go into implementing intensive techniques. Finally, we examined how our three models of decreasing accessibilities – Llama, GPT-OSS and Claude, varied in their performance on frame detection (**RQ3b**). On aggregate, Claude, the only proprietary model in our study, outperformed all the other language models. While GPT-OSS significantly trailed behind Claude, it showed somewhat of an improvement over both Llama and the discriminative models (**Table 3**).

# Discussion

Chiming into the evolving debate about computational methods for frame analysis, our study systematically evaluates three broad approaches to determine media frames in news articles. To test these approaches in a truly *in-situ* setting without benchmarks, we developed and put forth our own gold standard frames codebook and dataset, utilizing six months of news coverage of the US Mpox epidemic of 2022 as a case study. Specifically, we compared generative large language models against some of their popular computational predecessors (discriminative language models) and a traditional manual coding procedure. Not only do these approaches represent prominent chronological markers in the development of frame analytical methods, but they also vary vastly in their scalability, explainability and accessibility, thus allowing us to holistically assess the implications of our evaluations for frame analysis research. We performed this evaluation by investigating the utility of these approaches in three different tasks in our workflow: relevance classification – a processing step at the outset; frames codebook development – where we inductively identified developed an issue-specific frames typology from our corpus; and frame detection – where we deductively code articles for the presence of frames via a codebook-based content analysis. The variation of complexity and epistemic goals in these tasks additionally enabled us to navigate the classical tradeoff between automation and interpretation in computational social sciences (Jünger et al., 2022), and thus leverage the complementarity between different frame analytical approaches. Based on our findings, we present an updated roadmap on the best practices for news frame analysis (**figure 2**), summarizing our recommendations on how different approaches can be used in tandem to ensure solutions that favor reliability, without compromising on validity.

## Generative LLMs may complement, but not replace existing techniques

As foreshadowed by various instantiations about the difficulties of language models to grasp hermeneutical meaning (Choi et al., 2023; Henrickson & Meroño-Peñuela, 2025; Pinell, 2024; Sravanthi et al., 2024), the finding that LLMs performed worse than human coders was accentuated for particularly subjective tasks such as codebook development. Generally speaking, in line with previous work (Brown et al., 2025), when the tasks involved more intersubjective contestation (indicated by poorer agreement between humans), LLMs also struggled to achieve agreement with the gold standard. However, subjective differences and constructive disagreements in the interpretation of texts is a part of the qualitative research process. The iterative and self-reflexive components of our content analysis was designed to allow for a co-ordination of such diverging perspectives between different annotators. As demonstrated by the consistent improvement of reliability scores following the discussion of inter-coder disagreements between coding rounds (**Table 1, 3, 4**), intersubjective dialogue was key to achieving satisfactory outcomes with manual coding. Unfortunately, despite efforts to emulate a similar standpoint alignment by providing explicit human feedback to LLMs about their outputs (refer to Appendix sections A2.3.4 and A3.3.4), we did not see a significant improvement in performance. We argue that this failure to facilitate productive intersubjective dialogue and co-ordinate diverging interpretations between LLMs and humans was the key reason to their underperformance on complex, subjective tasks. Thus, since manual coding was the only approach in our study that consistently yielded substantial reliability scores, we recommend that frame analytical studies include a form of human validation from the outset. Nonetheless, we acknowledge that training research assistants is both time-consuming and expensive, and this process does not scale to corpora of sizes that are characteristic of media research today. Extrapolation with some sort of modeling is, in almost all cases, a necessity for downstream analyses. Finding tractable, defensible solutions to labeling such large-scale corpora is an ongoing conversation in the field.

Such extrapolation to a larger corpus is where we found the most promise for generative LLMs. Moreover, intersubjective dialogue and co-ordination of perspectives was not a requisite for every task; we found that for relatively simpler tasks, generative LLMs performed reasonably well. Examples of such tasks include relevance classification, and the detection of relatively straightforward frames[34]. However, we note two caveats: first, the task simplicity was not determinable a priori, since some form of manual coding on a small subset of articles was necessary to ascertain the level of intersubjective agreement, and by extension, task simplicity. Second, discriminative models often offered a more explainable and accessible solution for simpler tasks without compromising on reliability. In fact, for classes with high frequency of occurrence, they outperformed generative LLMs. For example, "Sexual stigma and transmission routes", the most frequent frame in our analysis, was most effectively detected by DeBERTa. In addition, when the classes could be roughly defined via lexical patterns (e.g., relevance), the Naïve Bayes classifier outdid all the generative models. However, we also acknowledge that the zero-shot capabilities of generative AI is compelling, since it bypasses the need for training data with a marginal loss in reliability. Generative LLMs then, offer a suitable approach for simple frame analytical tasks if one of the following conditions is met: the class frequency within the training dataset is low; or if the researchers cannot afford to perform manual coding to generate data to train their discriminative models.

With regards to implementation strategies, we found that, fine-tuning generative LLMs did not lead to a similar improvement in performance as training discriminative models typically does. This difference could be attributed to two reasons: training discriminative models involves a modification of the entire model, whereas fine-tuning generative LLMs only modifies the last few layers of the model (Parthasarathy et al., 2024). Research has also shown that where discriminative model performances plateau within a few thousand training examples (Burscher et al., 2014), generative LLMs show substantial performance increases going all the way up to a hundred thousand examples (Vieira et al., 2024). But generating fine-tuning datasets of such scale is infeasible for manual coders within a single study. Moving on, while a more exhaustive exploration of implementation strategies is beyond the scope of this paper, we explored a few other popular strategies (refer to Appendix section A2.3.4): such as utilizing LLM-as-a-judge and prompting for chain of thought (Farjam et al., 2025), indicator-based frame coding (Burscher et al., 2014) and verbalized confidence scores (Kwon et al., 2025; Yang et al., 2024). Although these approaches did lead to some improvements in the recall scores (i.e., fewer false negatives), they generally led to similar overall performances as the vanilla zero-shot strategy (refer to Appendix section A3.3.4).

---

[34] B (Racial disparities and stigmatizing name), D (Public health failure) and F (Human interest stories)

Overall, Claude was the only model in our study that yielded substantial outcomes ($\kappa > 0.6$). Two observations are worth noting here: first, the competitive edge that Claude maintained over the open-source models stemmed from its ability to better follow the codebook instructions and produce a balanced F1 score. Both Llama and GPT-OSS produced an excessive amount of type 1 errors (i.e., false positives), leading to precision scores that trailed even behind the encoder-only transformers. This error could be attributed to various reasons; generative LLMs have been known to exhibit label bias (Reif & Schwartz, 2024), and as such, the models may be biased to output "yes" over "no" when asked to determine if a frame is present in an article. Moreover, the models may have also struggled to follow the assigned codebook diligently when detecting frames, a problem that is also observed within manual coding research as coders experience more fatigue when annotating with long documents and codebooks. Indeed, prior research has proven the effectiveness of summarizing longer documents into smaller units before detecting frames (Kwon et al., 2025). Although our input texts were well within the token limit, models may currently lack the ability to follow verbose instructions[35], and thus could be inordinately liberal when marking for the presence of frames. However, reducing the input text length via indicator-based frame coding also failed to improve the overall performance (refer Appendix section A3.3.4). Second, while GPT-OSS trailed behind Claude, when the latter achieved substantial to near perfect agreement with the gold standard (e.g., frames B, C, F), so did the former. Relatedly, when GPT-OSS did not achieve substantial outcomes, neither did Claude. Therefore, larger, proprietary models could be used to improve performance on tasks where open-source models are already performing decently, but not otherwise. The question of whether to use proprietary models then, is more a question of accessibility and affordance, than that of reliability.

It is worth noting that the performance differences observed among the evaluated LLMs may partly stem from their distinct training paradigms. Claude employs Constitutional AI (Bai et al., 2022), which trains models against a set of predefined principles with minimal human oversight, whereas GPT (Ouyang et al., 2022) and Llama (Touvron et al., 2023) rely on Reinforcement Learning from Human Feedback (RLHF), which optimizes models based on human preference data. Constitutional AI's emphasis on consistency and principle-based reasoning may contribute to Claude's superior performance in our frame detection task, particularly for nuanced frames with lower prevalence. However, both training approaches face the same fundamental tension between alignment guardrails and the sensitivity required for value-laden analysis (Ji et al., 2025). Future research may systematically investigate how these training paradigms impact performance across different frame analysis tasks and whether certain approaches are better suited for interpretive social science research.

### Guidelines and limitations

Figure 2 summarizes our recommendations for how to perform news frame analysis in the face of the new computational developments, based on the results presented in this paper. Smaller datasets (few hundred articles) are best annotated via manual coding. For larger datasets, we suggest manually annotating a representative subset (50-100 articles), which could later on be used as a test set for validating the models that extrapolate onto the larger corpus of articles. Then, task simplicity can be determined by assessing if manual coding yields substantial outcomes ($\kappa > 0.6$) within the first iteration. If they do, then one may reliably use individual computational models for extrapolation. High frequency (>30%) classes are best annotated via supervised discriminative models; with smaller BoW models offering explainable solutions for simpler classes that can be grounded lexically (e.g. relevance), and larger, encoder-only transformers providing reliable outcomes for more complex classes. However, these supervised approaches are contingent on the researcher's ability to teach research assistants to generate training data spanning a few hundred articles. In cases where such an affordance is a barrier, or when the class frequency is relatively low (<30%), we recommend using generative (decoder-only) LLMs. Here, the choice of whether to use proprietary or open-source models may then be contingent on the researcher's ethics, affordances and privacy concerns. While using proprietary models may provide the most reliable outcomes, open-source options would still yield quality annotations on simpler tasks. Finally, when the task isn't simple, as determined by unsatisfactory inter-rater scores ($\kappa < 0.6$) between manual annotators on the test set, we recommend adopting pluralistic approaches combining the strengths of multiple language models, while forewarning that the final annotations may only yield moderate agreement with the gold-standard. Here, by pluralistic approaches, we mean methods that combine the strength of multiple approaches, such as ensemble learning (Burscher et al., 2014; Polikar, 2012), where one uses an ensemble of classifiers in tandem for annotation, or more recent collaborative frameworks between humans and generative LLMs (Egami et al., 2024; Gao et al., 2024).

---

[35] Our codebook contains a "when not to use" clause for each frame that directs coders to be vary of common misconceptions that lead to false positives.

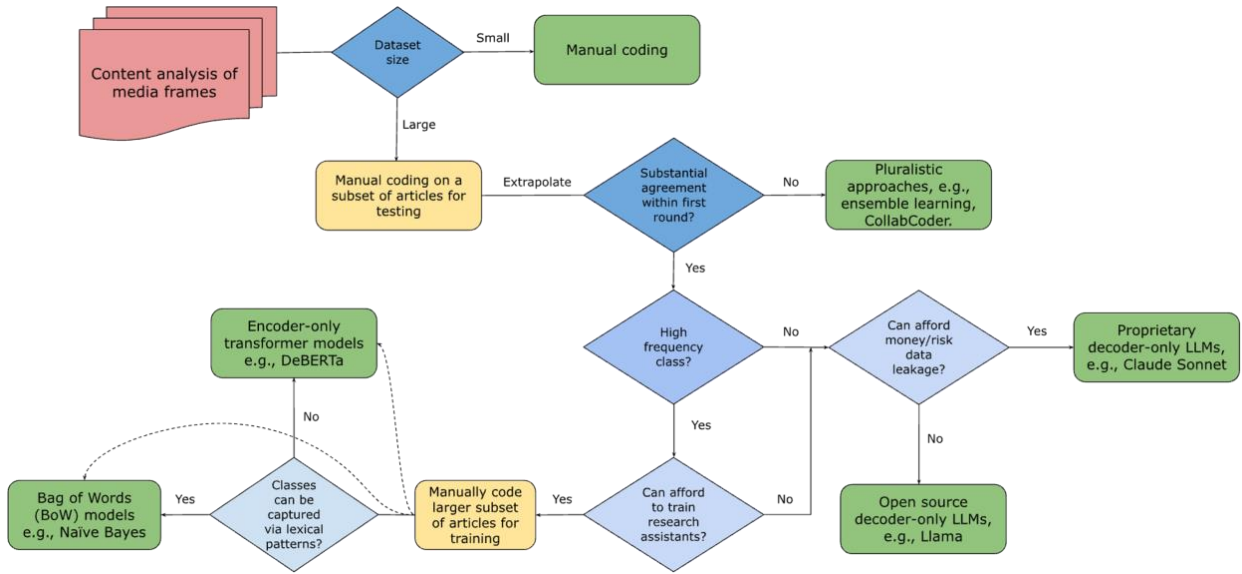

**Figure 2:** Guidelines for content analytical tasks within frame analysis: relevance classification and frame detection.

Note that our recommendations err on the side of accessible and explainable solutions, such that even when inaccessible options may provide an easier shortcut, we prioritize accessible options that do not compromise on reliability. We also acknowledge that the heuristics we use in our decision points (e.g. $\kappa > 0.6$ for simplicity) are operational, and hence maybe open to contestation based on the goals and standards of the researcher. In addition, these guidelines pertain to content analytical tasks within our study design (relevance classification and frame detection). For codebook development, given that scalability is lesser of a concern and computational approaches lack adequate hermeneutical depth, we urge scholars to adopt manual approaches on a representative sample. Generative LLMs could be used in a complementary capacity using the analytical design we outlined in Appendix section A2.3.3, but we restate the importance of human validation in this case.

Of course, our study is not without its own limitations. First and foremost, our analysis hinges on a particular case study of news frames surrounding a singular instance. In addition, while our operationalization of frames rests on previously documented best-practices (Entman, 1993; Matthes & Kohring, 2008), we acknowledge that bearing in mind the widely documented elusiveness of media frames, any far-reaching conclusions regarding frame analysis as a whole must be taken with a grain of salt. Although, in a landscape where LLM evaluations primarily rest on benchmark datasets[36] and deductive coding schemes, we hope that our case study stands out due to its *in-situ* nature. We offer a foundational design for a mixed-methods inductive frame analysis and invite future work to build upon it using similar case studies. Second, despite an already elaborate comparison, given the burgeoning state of language modelling, we are likely to have missed out on some potentially useful computational alternatives. We intentionally maintained basic implementation techniques, leaving out various error mitigation strategies. On a similar vein, while our comparative approach evaluated every method on its own, prior research has highlighted the potential of pluralistic approaches. Future work may explore these intensive options, and we hope that our novel manually validated dataset of relevance and frame labels provides a fruitful benchmark to test the plethora of methods to come. Third, by focusing only on English language corpora from news outlets within the United States, we reproduced a very prominent fallacy within computational text analysis research of constraining the scope of study to WEIRD[37] settings (Baden et al., 2022). In the wake of advances in multilingual LLMs (Qin et al., 2024), we echo a call to expand the scope of computational news frame analysis to non-WEIRD settings.

---

[36] That may already be present in the pre-training data

Finally, besides the many methodological caveats of computational approaches enlisted above, various ethical limitations within LLMs require deliberation. Unlike manual coding and supervised machine learning approaches where bias is limited to that of the researcher, biases within large language models depend on the pre-training data, mirroring biases within society at large (Brown et al., 2025). Research in AI alignment and bias mitigation attempts to lay guardrails on LLMs but generally tends to push them towards being excessively neutral (Rogers & Zhang, 2025), thus making them unsuitable for value-laden research. Additionally, the most powerful language model for frame detection in our study, Claude, is proprietary; researchers cannot audit these models effectively due to their obscurity of pre-training data, modelling architectures and training techniques (Ziems et al., 2024). As such, large language models also pose several privacy and accessibility risks. On one hand, while open-source models are free to use and more secure since they can be run locally, they demand reasonably high computational resources and literacy (Baden et al., 2022). On the other hand, although proprietary models run on their companies server's with more streamlined user interfaces, researchers are now faced with privacy risks, where their data may leak into the models during use, and accessibility constraints, since these models are expensive (Ziems et al., 2024). Notably, the data centers that house these models are in turn resource-intensive and currently constitute environmental hazards (Cisar, 2024), further compounding onto the ethical dilemma faced by researchers engaging in computationally intensive approaches. Although we limited the scope of our study to methodological aspects of frame analysis, it is equally, if not more crucial to be mindful of many such ethical challenges.

## Conclusion

Given the rapidly evolving state of generative AI and computational methods, it is also worth contemplating on the robustness of our findings over time. In our study, bigger models did yield better outcomes for complex tasks, indicating that as models get bigger in the future, we may observe transformative improvements in their performances. Innovations in multimodal models[38], AI alignment and pluralistic approaches may yet again lead to methodological advances. As such, one may ask: could AI-driven frame analysis yield performances that are satisfactory, and maybe even better than manual coders someday? Our response to this question is three-pronged: first, while more sophisticated models may get better at modelling language, their reliance on specific epistemic assumptions[39] may pose a fundamental limit on their capabilities (Pinell, 2024; Resnik, 2025). Frame analytical tasks often appeal to hermeneutical ideals, and by design lead to intersubjective disagreements that require co-ordination between different annotators. Even if such value-laden co-ordination of diverging standpoints is theoretically plausible[40] for LLMs, a lack of comprehension about their standpoints (Messeri & Crockett, 2024) and their appeals to a 'view from nowhere' (Mollema, 2025) pose significant hurdles. In other words, our recommendation regarding a necessity for manual validation and scrutiny over model outputs is likely to stand the test of time. Second, the superiority of larger,

---

[37] Western, Educated, Industrialized, Rich, and Democratic populations

[38] Models that take data of modalities beyond text (images, sound, videos etc.) as input

[39] Namely, the distributional semantics hypothesis, which posits that words that occur in similar contexts carry similar meanings. This hypothesis allows for models to 'understand' and approximate human-like speech by encoding statistical distributions of word/token co-occurrences.

[40] We were unable to find effective ways to facilitate such intersubjective co-ordination

proprietary models is concerning – as LLM size scales up, it provides diminishing returns in terms of performance (Hackenburg et al., 2025). Thus, even if we achieve human-like performances on complex tasks with LLMs, the models are likely to become too large and expensive for an average researcher to deploy. Accessible and open-source solutions are key to rigorous scientific research; finding feasible and practical alternatives is an urgent need of the hour. Third, frankly, it is next to impossible to predict where we go next in an ever-changing landscape of innovations in language modelling. An intellectually humble response to future questions about the capabilities of LLMs requires that we constantly monitor and validate models. We hope that our design framework and manually validated dataset provides opportunities for such testing and urge that researchers maintain a constructive yet critical attitude towards the application of black-box techniques in computational communication research.

To conclude, in the wake of generative AI, our study revisits computational frame analysis to present a comprehensive and updated evaluation of frame analytical tools. Specifically, utilizing the news coverage of the US Mpox epidemic of 2022 as a novel case study, we examined generative large language models against discriminative language models and manual coding across three different tasks that constitute a typical inductive frame analysis. Our results underscored manual coding as the reigning contender, with its utility also extending to the provision of guidance as to what and where computational methods could be effective. Overall, while computational approaches do provide feasible solutions to scale frame analysis to larger corpora, their performances were highly variable and needed careful human validation prior to application. Contrary to conventional wisdom, simpler and smaller models prevailed over heftier generative models in some cases, highlighting the need for a more pluralistic and comprehensive consideration of analytical methods when prioritizing for reliability. We culminate our study with a discussion of the ethical and methodological implications of our findings and summarize our recommendations with concrete frame analysis guidelines for future research.

## Acknowledgements

SVS acknowledges funding from the Patrick J. McGovern Foundation. The views and findings are those of the author and not those of Patrick J. McGovern Foundation. A.H.S. is supported by the National Science Foundation Graduate Research Fellowship Program under Grant No. 1938052. Any opinions, findings, and conclusions or recommendations expressed in this material are those of the authors and do not necessarily reflect the views of the National Science Foundation. Last but not least, we thank Claire Coffman for her contributions to the pilot data annotation efforts and other members of the Communication, Media and Marginalization (CoMM) lab for their valuable feedback on the manuscript.

## Data and Code Availability Statement

Data and code for all the analysis conducted in the manuscript is available at: https://github.com/sharajk/computational-framing

# Computational frame analysis revisited: on LLMs for studying news coverage

### Supplementary material

Sharaj Kunjar      Alyssa Hasegawa Smith      Tyler Mckenzie

Rushali Mohbe      Samuel V Scarpino      Brooke Foucault Welles

## A1. Data

### A1.1. Data collection

The raw data was collected using the MediaCloud webAPI [1]. We queried for articles containing the terms "mpox" or "monkeypox", published between May 1 to October 31, 2022 (6 months) by 248 national level news outlets in the US. This dataset consists of 7721 articles.

Then, texts, keywords and authors of these articles were webscraped using newspaper4k [2].

### A1.2. Pre-processing

#### A1.2.1 Selection

Here, articles were chosen for further analysis through the following steps:

1. Articles whose texts could not be scraped from newspaper4k were excluded.
2. Articles that had fewer than 250 words were excluded.
3. Articles that contain more characters than the excel cell limit (32767) are excluded.
4. Articles whose main texts didn't contain the term "monkeypox" or "mpox" were excluded.
5. Articles that contained the terms: "associated press", "wired" or "reuters" were removed to remove syndicated texts and maximize diversity in the dataset.

#### A1.2.2 Deduplication

Many articles are copy-pasted by the same outlet on different days or similar outlets. These articles should be included only once in the manual annotation stage to maximize variation within annotated articles.

We constructed a graph where nodes were articles and two articles A and B were connected by a directed edge (A, B) whose weight was equal to the fraction of words in A that were also seen in B. So if one article is "Whales are very cool and dolphins are also great" and the other article is "I think whales are great and they are also large", the edge weights would be 6/9 and 6/10 (from A to B and B to A) ("whales", "are", "and", "are", "also", "great").

Then we pruned the graph to only include edges whose weight was greater than or equal to a selection of thresholds ([0.75, 0.85, 0.9, 0.95, 0.975]). For each of these graphs, we pulled out all the strongly connected components and, for each SCC, we removed all but one article (the first one in the component detected by networkx, so this is fairly arbitrary) from the dataset.

We used the dataset with 95% threshold for assigning articles for manual annotation and training language models. Towards the end, when we are predicting annotations over articles that haven't been annotated, these deduplicated articles will be kept in the test set.

### A1.2.3 Boiler-plate removal

A lot of articles contain boiler plate texts (advertisements, information about the author, highlights/key takeaways etc.) that aren't relevant to the article's content. These texts need to be removed in order to optimise the language models.

To remove boilerplate, we iteratively examined articles from the same outlet, noting common phrases, scraping artifacts ("Advertisement Here"), and paragraphs that were repeated before/after the article text. We added such pieces of text to one of three lists: clip_before, noting chunks of text that came after various advertisements or other irrelevant information and directly preceded article text; clip_after, which kept track of chunks of text after which no article text would appear; and clip_at, noting scraping artifacts and other boilerplate text that appears within an article but is not part of the article text. We then removed text from each article, clipping anything that came before a chunk of text in clip_before (inclusive of the text in clip_before), anything that came after a chunk of text in clip_after, inclusive of the text chunk itself, and simply deleting anything in clip_at.

> After collection preprocessing, the curated dataset contains 3314 articles from 131 unique national level news outlets, ranging over a 6 month timeline of May-October 2022.

# A2.  Methods

## A2.1.  Relevance classification

Since our data collection was based on string matching for the term "monkeypox" or "mpox", our current dataset contains articles that could be about other issues, but only briefly mention mpox. Hence at this stage, we further processed articles to restrict our analysis to articles that were actually about Mpox. To this end, first, we perform a codebook-assisted qualitative analysis, as follows.

### A2.1.1 Human-generated labels

Codebook-assisted qualitative analysis was performed by the first and the third authors of the paper on a subset of the preprocessed articles, as follows.

- **Data familiarization:** The manual annotators spent a week going over a random subset of the articles (100/4142) and identified points of relevance/irrelevance within the subset. They met and discussed after going over the articles independently.

- **Codebook development:** The annotators then developed a codebook (attached below) for relevance classification based on their discussions.

- **Manual annotation:** The manual annotators then classify a set of 500 articles as relevant/irrelevant based on the attached codebook. These articles were chosen by randomly sampling with an assigned probability for each article in the entire pre-processed dataset (3314 articles).

  - After the first round of independent coding, the annotators achieved a Cohen's $\kappa = 0.82$. They disagreed over 41 articles.

  - In the second round, the annotators independently recoded the 41 articles that they disagreed on, then achieving a Cohen's $\kappa = 0.97$.

> Relevance codebook
>
> **Code: Relevant (1)**
> **Definition:**
> An article is marked *Relevant (1)* if it contains explicit content directly discussing Mpox in relation to the human epidemic context.
> **Criteria (one or more must apply explicitly):**
>
> - Mpox case counts, geographic spread, infection rates.

- Mpox transmission patterns, e.g., discussion of how it spreads between people.

- Information about Mpox vaccines, availability, development, distribution.

- Identification or discussion of at-risk groups (e.g., MSM, immunocompromised individuals, anyone who engages in close contact with a patient).

- Descriptions/critiques of the public health response (e.g., health advisories, CDC/WHO statements, local health campaigns).

- Details about epidemiological characteristics of Mpox (e.g., symptoms, strains, zoonotic origins when tied to the outbreak).

- Discussion of media coverage or representation of Mpox in the news or public discourse.

- Government policy, funding, or international efforts directly targeting Mpox.

- Personal stories about people dealing with Mpox/stigma due to Mpox.

**Code: Irrelevant (0)**
**Definition:**
An article is marked *Irrelevant* if it does not substantively discuss Mpox in a way related to the epidemic, or mentions it only in passing, satirical, or non-human contexts.
**Criteria (one or more must apply/none of the relevance criteria applies):**

- Incomplete Scrapes
    - Article is cut off mid-sentence or has only a headline or metadata. (Keep articles where they're cut off from a paragraph or so, leaving the scraped data still coherent, no need to cross check with links)

- Non-English articles

- Briefings/Quick Hits/Highlight Memos
    - Summarized, non-substantive listings with no in-depth coverage.
    - E.g., "Top 5 health stories today: flu, Mpox, RSV…"

- General Discussion of Public Health/Diseases/Vaccines
    - Mpox only mentioned as one of many diseases in a broad context.
    - E.g., "Vaccination efforts continue for diseases such as flu, Mpox, and measles."

- Satirical or Throwaway Mentions
    - Mpox used as a joke, stereotype, or stigmatizing punchline.
    - E.g., "Avoid gay men or you'll get Mpox — just kidding, but be careful!"

- Animal/Primate/Zoonotic Research (without human epidemic connection)
    - Mpox mentioned in the context of animal studies or general zoonoses.
    - E.g., "Study finds new poxvirus in African rodents; Mpox cited as historical example."

- Hyperlinked References Only
    - Mpox is only mentioned in a hyperlink or suggested reading, not in the body.
    - E.g., "…a variety of poxviruses [read more here]."

- General LGBTQ+ Issue Coverage
    - Mpox name-dropped in a list of issues affecting the queer community but not discussed.
    - E.g., "From mental health to Mpox, queer health is under threat."

**Note**

- Ambiguous Mentions: If it's unclear whether the article is relevant, lean toward exclusion and mark it *irrelevant*.

- Mixed Content: If an article contains some irrelevant framing (e.g., satire, general disease mentions) but also includes relevant details (case counts, public health response, etc.), mark as *Relevant*.

- Contextual Zoonosis: If zoonotic origins or animal vectors are discussed with direct connection to the current human outbreak, mark *Relevant*.

- Focus on explicit content — assume implicit relevance (e.g., inferred risk or allusion) is not sufficient.

### A2.1.2 Discriminative language models

In this section, we evaluated two transformer-based models, deberta-v3-base [3] and bert-base-cased [4], as well as a Naïve Bayes model that was trained using a bag-of-words approach. All models were trained on the article title combined with the article text. We trained the neural network binary classifiers using 200 articles in the training set, 150 in the validation set, and 150 in the test set. The Naïve Bayes models were trained using a 300/200 train/test split. We selected the transformer-based models for their widespread adoption and because we could fit one model per frame on a 32 GB NVIDIA Tesla V100 SXM2 GPU; the Naïve Bayes model was selected as a simple baseline comparison. We trained both transformer models for 20 epochs, evaluating every 25 training steps, with two articles per training batch. Training for fewer epochs frequently led to a zero F1 score on the validation data and subsequent poor performance on the test set. We chose the saved model for each frame that had the best F1 score on the validation set during training as that frame's final model. Both models performed best on the validation data with a learning rate of 5e-5. A computing instance with 8 GB of RAM and two CPU cores was sufficient to run this training along with the aforementioned GPU.

### A2.1.3 Generative large language models

Here, we evaluate a few different large language models: Llama-3.1-8B Instruct (8-bit quantized) [5], Claude Sonnet 4 [6], and GPT-OSS-20B [7]. The Llama models were run on a local machine with a 16GB Apple M1 chip using an open-source application Ollama [8]. The Claude models were run on Anthropic's servers using their Python API client [9]. The GPT-OSS models were run on a 140GB NVIDIA H200 GPU.

Due to their high computing requirements, the Claude and GPT models were only tested in the zero-shot setting. The llama models were tested in both zero-shot and fine-tuned settings. All the zero-shot tests were done on the entire set of 500 manually labelled articles. The fine-tuned settings were performed with a 300/200 train/test split. The prompts used for annotations are attached below:

---

LLM prompts for relevance classification

**SYSTEM_PROMPT =**"You're a communication researcher who is studying the news reporting of Mpox. You'll perform relevance coding on news articles, by classifying articles as relevant, or not relevant. Remember to prioritize accuracy and clarity in your analysis, using the provided context and your expertise to guide your evaluation. Here's the codebook for you to follow for class definitions: " + RELEVANCE_CODEBOOK **USER_PROMPT =**"Is the following article relevant? Provide your response in a JSON array format, as follows, and include nothing else in the response:{"relevance": "yes/no"}. If you are uncertain about the classification, force a decision to choose "yes" or "no". " + ARTICLE_TEXT

---

After relevance classification, we obtained a set of 2224 relevant articles (as extrapolated using BERT). All further analysis will be based on these articles.

## A2.2. Codebook development

In this section we developed a issue-specific frame detection codebook using a random subset of a 100 (out of 2224) relevant articles. We tested and evaluated the codebooks developed using 3 techniques: applied thematic analysis (manual), transformers-based topic modelling, and LLM-based inductive thematic analysis.

### A2.2.1 Human-generated codebook

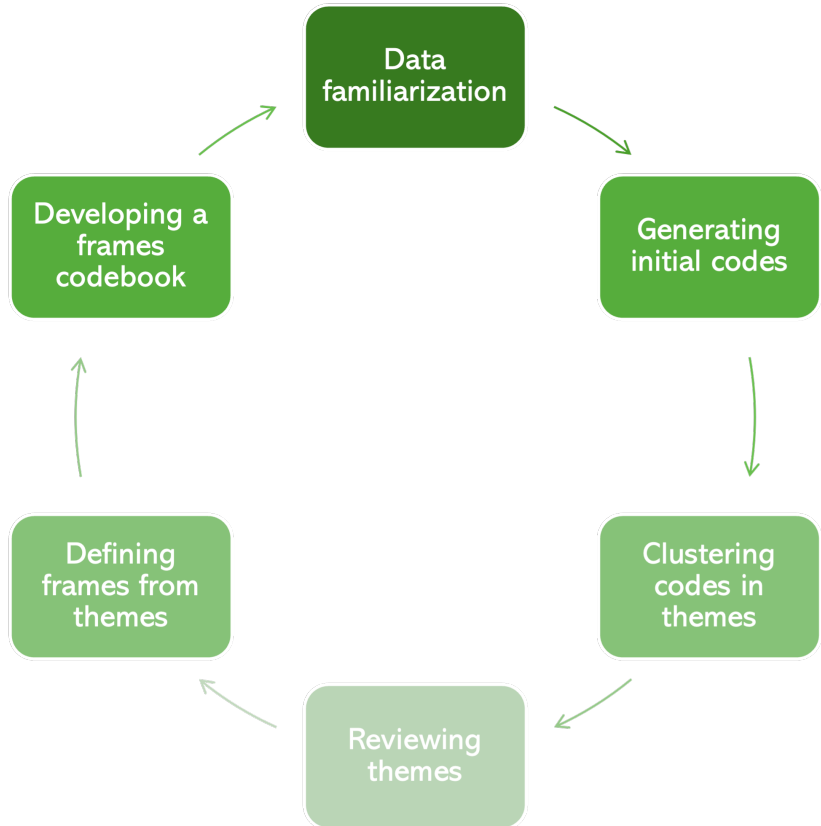

Figure A1: Six-phase analytical procedure for developing a frames codebook, adapted from Braun and Clarke's thematic analysis [10].

The first and the third authors of the paper performed applied thematic analysis [10–12] on 100 articles, with the following methodological and epistemic foundations:

- We treated each article as the unit of analysis. Consistent with the rest of the study, we decided to treat each article as representative of the narrative frame that the author of the article intends to communicate.

- Our analysis was **inductive**, i.e., did not rely on a pre-existing typology (ergo deductive) of framing, since issue-specific frames tend to be contextually specific and aren't appropriately pre-defined by the existing literature for our use case.

- Our analysis was **interpretive**, i.e, we urged the researchers to go beyond explicit lexical cues and denotations when interpreting the texts. While interpretive research does forego an appeal to value-detached objectivity due its inherent subjectivity, we maintain intersubjectivity [13] by evaluating the reliability of the generated codebook via inter-rater scores.

- Our analysis was **iterative** and **reflexive**. The manual coders developed their codebooks over two rounds of analysis. While the annotations within each round was independent between coders, they were encouraged to discuss their outcomes after each round, and reflect upon how their standpoint on the articles affected their interpretation of the same.

- We follow Matthes and Köhring's [14] procedure to realize and define frames in terms of **Entman's frame elements -** problem definitions, causal interpretations, moral evaluations and treatment recommendations [15].

Further details on the procedure invoked to develop the codebook will be elucidated in later work (Kunjar, forthcoming).

## A2.2.2  Discriminative language models

We implemented BERTopic [16] on the entire set of 2224 relevant articles after cleaning the texts for stop words.

## A2.2.3  Generative large language models

We followed instructions provided by Dai 2023 [17] and De Paoli [18] to perform inductive thematic analysis using LLMs. Due to its low compute requirements, we used Llama-3.1-8B-Instruct (8-bit quantised) for this task. The prompts used for this task are attached below. The procedure broadly involved three steps:

1. **Generating initial codes:** Here, we provided the model with a set of 100 articles (one at a time) and prompted it to generate codes for each article. The model generated 213 unique codes at this point.
2. **Clustering codes into themes:** Here, we curate all the initial codes generated by the model and prompt it to cluster them based on conceptual similarity. The model produced 21 themes.
3. **Reviewing themes and defining frames:** Here, we finally review the themes produced by the model and prompt the model to create generate a set of unique frames based on Entman's [15] definition of framing.

---

**Prompts for codebook development with Llama**

**Generating initial codes**

```
SYSTEM_PROMPT ="You're a communication researcher who is studying the news coverage
of the Mpox epidemic. You will be performing inductive thematic analysis. Your job
to generate initial codes for articles. Your expertise is crucial in identifying
issue-specific frames that can sway pubic opinions and distort public discourse.
Avoid generic or overly broad categories. Do not assume predefined categories, it is
imperative that the codes are driven by the data. Think carefully to determine codes
that are representative of the articles but also have maximum variability between
them.
Format your response has to be a JSON array of objects, and include nothing else in
the response. Each object should contain the following keys: { 1. "Index": "A number
indicating the code order", 2. "Name": "A short name for the code (max 3 words)", 3.
"Description": "A 3-line explanation of the code", 4. "Example": "A representative
quote (max 2 sentences)".}"
USER_PROMPT ="Attached below is an article about Mpox. Identify upto 3 unique codes
and prevelant codes from the text. Ensure that your output format adheres to the
instructions provided." + ARTICLE_TEXT
```

---

**Clustering codes into themes**

```
SYSTEM_PROMPT ="You're a communication researcher who is conducting an
issue-specific narrative frame analysis on the news coverage of the Mpox epidemic. A
researcher has already gone over news articles and come up with a list of initial
codes through inductive coding. Your job is go over a set of initial codes and
cluster them to flesh out unique frames. In his seminal work, Entman said "Framing
essentially involves selection and salience. To frame is to select some aspects of
a perceived reality and make them more salient in a communicating text, in such a
way as to promote a particular problem definition, causal interpretation, moral
evaluation, and/or treatment recommendation for the item described." Follow Entman's
principles and define frames based on the 4 unique framing elements. Prioritize
frame clarity and uniqueness.
Format your response as an array of upto 3 JSON objects, and include nothing else
in the response. Each object should represent a frame as follows: { "Name": "A short
```

```
name for the frame (max 3 words)", "Problem definition": "How do this frame define
the problem at hand?", "Causal attribution": "What actors or forces does this frame
attribute the cause of the problem to?", "Moral evaluation": "What value judgements
are being made by this frame?" "Treatment recommendation": "What remedies does the
frame suggest for tackling the problem?" "Example": "A representative quotes (max 2
lines each)". }"
USER_PROMPT ="Ensure that you adhere to the output format provided. Go over the
attached codes and group them into frames for a frame analysis. The following text
contains all the identified codes as a JSON array of objects." + INITIAL_CODES
```

---

### Reviewing themes and defining frames

```
SYSTEM_PROMPT ="You're a communication researcher who is conducting an
issue-specific narrative frame analysis on the news coverage of the Mpox epidemic.
In his seminal work, Entman said "Framing essentially involves selection and
salience. To frame is to select some aspects of aperceived reality and make them
more salient in a communicating text, in such a way as to promote a particular
problem definition, causal interpretation, moral evaluation, and/or treatment
recommendation for the item described." A researcher has already gone over news
articles and come up with a list of frames with Entman's framing elements. Your job
is to go over all the frames, group them based on conceptual similarity and produce
a revised set of frames with upto 10 unique frames. Follow Entman's principles
and define frames based on the 4 unique framing elements. Take your time to think
carefully and prioritize frame clarity and uniqueness.
Format your response as an array of upto 10 JSON objects, and include nothing else
in the response. Each object should represent a frame as follows: { "Name": "A short
name for the frame (max 3 words)", "Description": "2-3 lines of detailed description
of the frame", "Problem definition": "How do this frame define the problem at
hand?", "Causal attribution": "What actors or forces does this frame attribute the
cause of the problem to?", "Moral evaluation": "What value judgements are being made
by this frame?" "Treatment recommendation": "What remedies does the frame suggest
for tackling the problem?" "Example": "A representative quotes (max 2 lines each)".
}"
USER_PROMPT ="Go over the attached set of frames and produce a revised set of upto
10 unique frames for a codebook based annotation procedure. The following text
contains all the identified frames as a JSON array of objects." + CLUSTERED_THEMES
```

## A2.3. Frame detection

### A2.3.1 Human-generated labels

The first and the third authors performed a codebook-assisted annotation procedure (codebook attached in results) on a subset of relevant articles. The annotation was one-to-many, i.e., for each article, they simultaneously detected for all seven frames.

- The manual coders annotated a set of 100 articles (sampled at random) over two rounds.

  - They achieved an average Cohen's $\kappa = 0.57$ in the first round.

  - They discussed their disagreements and independently recoded the articles that they disagreed to achieve a Cohen's $\kappa = 0.84$.

- The first then proceeded to annotate an additional subset of 400 articles for training language models in the subsequent analysis.

### A2.3.2 Discriminative language models

Here, we used the same 3 discriminative language models - Naïve Bayes, BERT-base-cased and DeBERTa-v3-base to perform frame detection (using the same computing resources) as mentioned in the relevance classification section. All the models were trained on the labels produced by the first author on 400 articles (60/40 train/eval split), and tested on the 100 articles that were doubly annotated by both manual annotators.

### A2.3.3 Generative large language models

Here, we evaluated the same models (with the same settings and computing resources) as mentioned in the relevance classification section. All the zero-shot tests were done on the set of 100 articles that were manually labelled by both annotators. The fine-tuned was performed using the remaining 400 articles manually labelled by the first author. The prompts used for annotations are attached below:

---

LLM prompts for frame detection

```
SYSTEM_PROMPT ="You're a communication researcher who is studying the news reporting
of Mpox. You'll perform a codebook assisted frame analysis on news articles.
Identify the frames from the codebook that are present in the article. Here is the
codebook with frame definitions: " + FRAMES_CODEBOOK
USER_PROMPT ="Identify frames in the article with the following guidelines:
1. Read the entire article carefully before coding
2. Identify all the frames present in each article
3. Some articles maybe irrelevant to the Mpox epidemic or may not contain any of the
frames mentioned in the codebook. In this case, mark "no" for every frame.
4. Ensure at least 2 framing dimensions are explicitly present (unless noted
otherwise)
5. Use frame descriptions and examples to guide decision
6. All coding should be based on EXPLICIT presence of the frame. Thus, if a frame
isn't explicitly present, but implicitly implied, than mark "no" for that frame.
Provide your response in a JSON array format, as follows, and include nothing else
in the response:
{"sexual stigma and transmission routes": "yes/no",
"racial disparities and stigmatising name": "yes/no",
"global relations": "yes/no",
"public health failure": "yes/no",
"epidemic preparedness and surveillance": "yes/no",
"human-interest stories": "yes/no",
"broader health issues": "yes/no"}.
If you are uncertain about the annotations for any frame, force a decision to choose
"yes" or "no". "+ ARTICLE_TEXT
```

---

### A2.3.4 Error analysis and mitigation

Following error analysis, we attempted various error mitigation strategies to improve frame detection outcomes. Again, due to its ease of use and open-source nature, we opted to run these analyses with llama-3.1-8B-instruct (8-bit quantized). All prompts used are attached below.

- **LLM as a judge:** Hallucinations, stochasticity and complex instructions remained a barrier to reliable frame detection outcomes. Following the suit of agentic workflow put forth by Farjam et al. [19], we attempted to improve LLM outcomes by decomposing the task into two steps. First, we prompted two LLM agents (classifiers) to annotate the train set with the frame labels while providing a chain-of-thought reasoning. Second, we prompted another LLM agent (judge) to evaluate the responses provided by the classifier and output the final predictions. Using a chain-of-thought reasoning elicits better reasoning capabilities and improves transparency and interpretability. Pooling multiple answers (in our case, two) for the same input article allows for a more diverse set of predictions and thus creates room for the stochasticity that is inherent to LLM outcomes. Moreover, evaluating the responses using a third LLM (judge) improves reliability due to a potential to rectify hallucinations, and oversee adherence to the coding guidelines provided.

- **Human feedback:** The qualitative analytical procedures we employed were iterative in nature - where the coders annotated a set of articles independently, met to discuss disagreements, and then returned to annotate articles again. We attempted to mimic this approach with LLMs, so that they could align themselves better with the manual coders. This technique was inspired by RHLF (reinforcement learning with human feedback), where humans provide feedback to LLM responses to align model predictions with human preferences before the final prediction task. First, we prompted the LLM to provide a frame labels and chain-of-thought reasoning for a set of 30 articles (chosen randomly from the 400 articles in the train set). Second, the first author evaluated each prediction manually and provided a response regarding how the model performed, where it went wrong, and what the correct labels would be. Third, we fine-tuned the model with these exchanges. Finally, we prompt the fine-tuned model to provide predictions on the test set. Thus, this technique combines the chain-of-thought reasoning capabilities along with explicit human inputs and reasoning, allowing for a more interactive and iterative approach.

- **Decision trees:** Our coding task comprised of a reasonably long and detailed codebook. While the models we used allowed for long input token lengths, longer inputs and nested tasks often lead to worse outcomes. Here, we attempted to resolve this issue by decomposing our task to finest grain possible. First, we converted the codebook into a list of 55 yes/no questions based on the description of each frame and its frame elements (8 questions per frame on average). Second, we prompted to LLM to simply answer these questions, a few at a time, without any explicit reasoning. Then, we devised a decision-tree that deterministically assigns a label for each frame based on the responses to these questions. Thus, we were able to drastically downsize the input token length, and reduce some of the stochasticity due to the deterministic nature of the decision-tree.

- **Confidence scores:** One of the significant ways in which the iterative process improves qualitative coding is by a co-ordination of uncertainty. That is, the levels of certainty that different coders require in order to annotate a frame as "present" may vary. Consider this example: two coders A1 and A2 are 70% certain that a frame appears in an article X, but while coder A1 considers a frame F as present in article X if they're more than 60% certain, coder A2 considers a frame F as present in X only if they're more than 75% certain. Iterative coding procedures allows for a shared conception of this threshold. We attempted to circumvent this problem with LLMs by prompting the model to provide confidence scores instead of "yes/no" predictions. First, the model is prompted to provide confidence scores on a 100 articles (from the train set). Second, we binarised the predictions using a threshold for each frame, and estimate the threshold that maximizes Cohen's $\kappa$ in comparison to the human labels. Third, we prompted the LLM to annotate the test set with confidence scores for each article, and then deterministically binarize the scores based on the thresholds we estimated earlier.

- **Class Imbalance management for discriminative models:** The different classes in our analysis have varying levels of prevalence within the training data. Under-represented classes tend to do poorly in supervised algorithms. To manage this imbalance, we implemented focal loss [20], which is designed to down-weight easy negatives and up-weight difficult positives, potentially helping low frequency frames where positive evidence is scarce. We also learn different thresholds for different frames instead of using the default threshold of 0.5, so that the decision rules can be tuned for each frame.

---

LLM prompts for error mitigation

### LLM as a judge

### Classifiers

```
SYSTEM_PROMPT = "You're a communication researcher who is studying the news
reporting of Mpox. You'll perform a codebook assisted frame analysis on news
articles. Identify the frames from the codebook that are present in the article.
Here is the codebook with frame definitions:" + CODEBOOK
USER_PROMPT = "Identify frames in the article with the following guidelines:
1. Read the entire article carefully before coding
2. Identify all frames present in each article
3. Some articles maybe irrelevant to the Mpox epidemic or may not contain any of the
frames mentioned in the codebook. In this case, mark "no" for every frame.
4. Ensure at least 2 framing dimensions are explicitly present (unless noted
otherwise)
5. Use frame descriptions and examples to guide decisions
6. All coding should be based on EXPLICIT presence of the frame. Thus, if a frame
isn't explicitly present, but implicitly implied, than mark "no" for that frame.
```

Provide your response in a JSON array format with the answers and justifications, as follows, and include nothing else in the response:
[
{
"frame": "sexual stigma and transmission routes",
"presence": "yes/no",
"justification": "a detailed chain of thought explaining your decision",
"quote": "if the frame is present, an example quote from the article demonstrating it"
},
{
"frame": "racial disparities and stigmatising name",
"presence": "yes/no",
"justification": "a detailed chain of thought explaining your decision",
"quote": "if the frame is present, an example quote from the article demonstrating it"
},
{
"frame": "global relations",
"presence": "yes/no",
"justification": "a detailed chain of thought explaining your decision",
"quote": "if the frame is present, an example quote from the article demonstrating it"
},
{
"frame": "public health failure",
"presence": "yes/no",
"justification": "a detailed chain of thought explaining your decision",
"quote": "if the frame is present, an example quote from the article demonstrating it"
},
{
"frame": "epidemic preparedness and surveillance",
"presence": "yes/no",
"justification": "a detailed chain of thought explaining your decision",
"quote": "if the frame is present, an example quote from the article demonstrating it"
},
{
"frame": "human-interest stories",
"presence": "yes/no",
"justification": "a detailed chain of thought explaining your decision",
"quote": "if the frame is present, an example quote from the article demonstrating it"
},
{
"frame": "broader health issues",
"presence": "yes/no",
"justification": "a detailed chain of thought explaining your decision",
"quote": "if the frame is present, an example quote from the article demonstrating it"
}
].
If you are uncertain about the annotations for any frame, force a decision to choose "yes" or "no" and then report that uncertainty in your justification. "

**Judge**

SYSTEM_PROMPT = "You are a communication researcher who is studying news framing surrounding the Mpox epidemic. Two classifiers have read a newspaper article and identified whether certain frames from a codebook are present in the article or not. Your job is to go over their responses, chains of thought and the codebook yourself and make the final decision regarding whether the frames are present in the article. The classifiers tend to overlook the "explicit" clause in the codebook that mentions that any presence of frames must be based on explicit presence, so
ensure that you check for that. Remember to prioritize accuracy and clarity in your
analysis, using the provided context and your expertise to guide your evaluation.
Here is the original codebook used for analysis:" + CODEBOOK
USER_PROMPT = "Here are the responses from the two classifiers:
Classifier 1: CLASSIFIER_OUTPUT_1
Classifier 2: CLASSIFIER_OUTPUT_2
Provide your response in a JSON array format, as follows, and include nothing else
in the response:
{"sexual stigma and transmission routes": "yes/no",
"racial disparities and stigmatising name": "yes/no",
"global relations": "yes/no",
"public health failure": "yes/no",
"epidemic preparedness and surveillance": "yes/no",
"human-interest stories": "yes/no",
"broader health issues": "yes/no"}.
If you are uncertain about the annotations for any frame, force a decision to choose
"yes" or "no.""

---

**Human feedback**

**Round 1**
Prompts same as the classifier prompts in "LLM as a judge".

**Round 2**
Prompts same as mentioned in "LLM prompts for frame detection" in section 1.4.3.

---

**Decision trees**

SYSTEM_PROMPT = "You're a communication researcher who is studying the news
reporting of Mpox. You'll be analysing an article and answering a series of "yes
or no" questions. Here are the questions: " + DECISION_TREE_QUESTIONS
USER_PROMPT = "Ensure that you follow these guidelines:
1. Read the entire article carefully before answering
2. Some articles maybe irrelevant to the Mpox epidemic. In this case, mark "no" for
every question.
3. Answer each and every question that is asked.
4. All answers must be based on an EXPLICIT interpretation of the article. Thus, if
the answer for a question is "yes", but only based on what is implicitly implied,
mark "no" for that question.
5. If you are uncertain about the answer for any question, force a decision to
choose "yes" or "no".
Provide your response in a JSON array format, as follows, and include nothing else
in the response: { "question 1": "yes/no", "question 2": "yes/no".......}
Here's the article: " + ARTICLE_TEXT

---

**Confidence scores**

For each frame f we prompt with the following:
SYSTEM_PROMPT = "You're a communication researcher who is studying the news
reporting of Mpox. You'll perform a codebook assisted frame analysis on news
articles.
You're interested in detecting the presence of the frame: f. Here is the codebook
definition of the frame:" CODEBOOK_FOR_f
USER_PROMPT = Identify the probability that the frame ("""" + frame + """") is present
in the news article using the following guidelines:
1. Read the entire article carefully before coding
2. Some articles maybe irrelevant to the Mpox epidemic. In this case, mark '0'.

```
3. Ensure at least 2 framing dimensions are explicitly present (unless noted
otherwise)
4. Use frame description and examples to guide decisions
Provide your response in a JSON array format as follows, and include nothing else in
the response:
{"confidence": "A number between 0 (definitely absent) and 1 (definitely present)
indicating the probability that the frame exists in the article"}.
Here's the article:" + ARTICLE_TEXT
```

# A3. Results

## A3.1. Relevance classification

Table A1: Relevance classification metrics for all annotators. All language model predictions are evaluated against the labels produced by the first author after round 2 of manual annotation.

| Annotator | Accuracy | Kappa | F1 | Precision | Recall |
|---|---|---|---|---|---|
| Human (Round 1) | 0.92 | 0.82 | 0.94 | 0.97 | 0.91 |
| Human (Round 2) | 0.99 | 0.97 | 0.99 | 0.99 | 0.99 |
| Naive Bayes | 0.96 | 0.90 | 0.97 | 0.96 | 0.98 |
| BERT | 0.96 | 0.91 | 0.97 | 0.97 | 0.97 |
| DeBERTa | 0.98 | 0.95 | 0.98 | 0.99 | 0.98 |
| Llama zero-shot | 0.92 | 0.81 | 0.94 | 0.94 | 0.95 |
| Llama fine-tuned | 0.95 | 0.89 | 0.96 | 0.95 | 0.98 |
| Claude zero-shot | 0.95 | 0.88 | 0.96 | 0.94 | 0.99 |
| GPT-OSS zero-shot | 0.92 | 0.80 | 0.94 | 0.91 | 0.97 |

## A3.2. Codebook development

### A3.2.1 Human-generated codebook

After following the standard six-phase thematic analytical procedure [21], the manual coders came up with 7 frames, explicated as follows.

> **Frame detection codebook**
>
> **Coding Guidelines**
> 1. Read the entire article carefully before coding
> 2. Identify all frames present in each article
> 3. Mark articles as "irrelevant" (if the article is irrelevant to mpox coverage) or "no frame" (if the article is relevant to mpox but doesn't contain any of the below frames) when appropriate.
> 4. Ensure at least 2 framing dimensions are explicitly present (unless noted otherwise)
> 5. Use frame descriptions and examples to guide decisions
> 6. Record coding decisions and uncertainties systematically
>
> **Frame A: Sexual Stigma and Transmission Routes**
> - **Problem Definition:** Addresses whether Mpox disproportionately affects MSM communities and debates classification as STI or infectious disease spread through close contact
>
> - **Causal Interpretation:** Explains transmission through sexual contact/behaviors, general close physical contact, or combination of factors; attributes concentration in MSM communities to sexual practices, coincidental network effects, or other social factors
>
> - **Moral Evaluation:** Ranges from emphasizing stigma avoidance and dignity through neutral reporting to moral judgments about sexual behaviors and lifestyle choices

- **Treatment Recommendation:** Spans from stigma-informed public health messaging through targeted education to behavioral interventions targeting specific communities

- **Examples:**
  - "It's less about sexual identity and more about sexual networks"
  - "Health officials emphasize that while MSM communities are disproportionately affected, transmission occurs through close physical contact"
  - "Yes, Fauci, Monkeypox Is a 'Gay Disease' "

- **When not to use:** Simple epidemiological reporting without moral judgment; stigma discussions only about racial aspects of "monkeypox" name

## Frame B: Racial Disparities and Stigmatizing Name

- **Problem Definition:** Identifies racial disparities in Mpox cases, vaccine access, and healthcare treatment; debates whether "monkeypox" name is racially stigmatizing

- **Causal Interpretation:** Attributes disparities to systemic healthcare barriers and historical neglect, individual factors, or dismisses concerns as political correctness

- **Moral Evaluation:** Presents health equity as moral imperative, acknowledges concerns without strong position, or views racial stigmatization concerns as excessive or politically motivated

- **Treatment Recommendation:** Advocates for equitable distribution and virus renaming, maintains status quo approaches, or resists targeted interventions while favoring "color-blind" approaches

- **Examples:**
  - "Black and Hispanic people have been disproportionately affected by monkeypox—with disparities that are increasing over time—yet they are receiving a lower share of vaccines"
  - "The city is working to address vaccine access disparities while examining factors contributing to different infection rates across communities"
  - "Originally, we couldn't call monkeypox 'monkeypox' because that was somehow racist"

- **When not to use:** Race topics unrelated to Mpox; mere mention of racial demographics without discussing disparities

## Frame C: Global Relations

- **Problem Definition:** Frames Mpox as global health challenge requiring international cooperation, national issue with international implications, or internal affairs issue for individual nations

- **Causal Interpretation:** Attributes epidemic to wealthy nations' failure to help endemic regions, complex international factors, or failing healthcare systems in affected countries

- **Moral Evaluation:** Emphasizes global health equity and condemns wealthy nation neglect, acknowledges competing priorities, or prioritizes national security and citizen protection

- **Treatment Recommendation:** Advocates coordinated international response and vaccine sharing, balanced domestic and international approach, or travel restrictions and domestic priority

- **Examples:**
  - "Monkeypox has been a developing problem for decades and the current global outbreak was avoidable, but the looming threat was largely ignored"
  - "International cooperation will be essential for vaccine distribution while countries also address their domestic needs"
  - "There have been 72 deaths reported, but all in sub-Saharan Africa where health care, in technical terms, really sucks"

- **When not to use:** Simple mentions of international spread without taking globalist/isolationist position

**Frame D: Public Health Failure**
- **Problem Definition:** Criticizes public health response as slow, inadequate, or poorly coordinated; highlights institutional incompetence

- **Causal Interpretation:** Attributes control challenges to policy choices, institutional weaknesses, funding shortfalls, or historical disinvestment

- **Examples:**
  - "Across the nation, public health agencies are running too few tests, and lack the infrastructure to execute a coordinated response"
  - "Most Americans do not trust the Biden administration or Dr. Anthony Fauci to properly handle a monkeypox outbreak"

- **Note:** This frame is primarily diagnostic (one dimension sufficient). Focuses on evaluating public health infrastructure with explicit implications for governmental institutions.

- **When not to use:** Simple reporting of public health measures without evaluation; outbreak challenges without attributing to systemic failures

**Frame E: Epidemic Preparedness and Surveillance**
- **Problem Definition:** Presents established institutions as appropriate leaders, mixed assessment of institutional effectiveness, or frames institutions as ineffective/overreaching; characterizes response as under-reaction, proportionate, or alarmist over-reaction

- **Causal Interpretation:** Places responsibility on institutional systems, shared responsibility across multiple factors, or emphasizes individual and risk group responsibility; attributes success/failure to institutions, complex factors, or individual compliance

- **Moral Evaluation:** Emphasizes egalitarianism and urgent action necessity, balances collective and individual concerns, or prioritizes individual freedom and proportionate response

- **Treatment Recommendation:** Recommends declaring emergencies and increased surveillance, supports measured institutional response, or opposes emergencies and advocates deprioritization

- **Examples:**
  - "Last weekend, the WHO elevated monkeypox to a public health emergency of international concern (PHEIC). Public health experts say that it's a decision that should have come weeks ago"
  - "Officials are weighing the benefits of emergency declarations against concerns about public fatigue and resource allocation"
  - "Democrats may use monkeypox as yet another reason to lock us down and force mail-in voting upon us"

- **Note:** This frame is primarily prognostic (one dimension sufficient). Expresses trust/mistrust in government establishments.

- **When not to use:** Simple reporting of measures without advocating institutional trust/distrust; points covered by Frame D alone

**Frame F: Human-Interest Stories**
- **Problem Definition:** Presents personal experiences and episodic narratives from affected communities, emphasizing human side of epidemic

- **Treatment Recommendation:** Emphasizes community-based support, stigma reduction, healthcare access improvement, and centering affected perspectives

- **Examples:**
  - "On Friday, June 17, I got a call from a friend I'd been hanging out with the weekend prior, informing me that I had likely been exposed to monkeypox. For about five to seven days, it just felt like I had crazy abdominal cramps constantly"
  - "Local support groups have formed to help people navigate testing and treatment while dealing with stigma in their communities"

- "Lilly Simon, a 33-year-old in Brooklyn, does not have monkeypox. She does have neurofibromatosis type 1, a genetic condition that causes tumors to grow at her nerve endings. Those tumors were filmed surreptitiously by a TikTok user while Ms. Simon was riding the subway" (mistaken identity narrative)

- **Note:** This frame is episodic (one dimension sufficient). Focuses on personal narratives rather than structural analysis.

- **When not to use:** Brief mentions of individual cases without narrative development; personal accounts from officials/professionals.

**Frame G: Broader Health Issues**

- **Problem Definition:** Places Mpox within broader context of infectious diseases and systemic health issues, acknowledges connections to other health challenges, or emphasizes Mpox's distinctiveness while comparing to other pressing issues

- **Causal Interpretation:** Attributes multiple disease threats to shared structural causes requiring comprehensive solutions, recognizes multiple contributing factors, or emphasizes specific causes while critiquing institutional overreach

- **Moral Evaluation:** Emphasizes collective responsibility and systemic change imperative, balances individual and collective concerns, or prioritizes individual liberty and efficient resource targeting

- **Treatment Recommendation:** Advocates strengthened public health infrastructure and addressing root causes, supports targeted improvements, or focuses on pragmatic immediate threat management and fund diversion

- **Example:**

  - "We are in danger as a species of more infectious diseases, but we're not putting the public health infrastructure in place… America should be prepared. Not responsive, but proactive"
  - "Health experts say monkeypox is unlikely to wreak the same kind of havoc as Covid, which has killed millions, infected more than half a billion people, and ravaged the world's economy"
  - "Any of the politicians trying to scare you about this, do not listen to their nonsense. We're not going to go back to Fauci in the '80s trying to tell families they are going to catch AIDS by watching TV together"

- **Note:** This frame is thematic - requires significant deviation from Mpox focus (few sentences about other health issues sufficient).

- **When not to use:** Articles focusing strictly on Mpox; brief mentions of other diseases without substantive connections.

## A3.2.2 Discriminative language models

**BERT based topics**

Table A2: Top 10 topics as identified by a BERT topic model applied on all the relevant articles

| Topic | Name | Representation |
|-------|------|----------------|
| 1 | vaccine we doses about | ['vaccine', 'we', 'doses', 'about', 'were', 'more', 'vaccines', 'said', 'but', 'you'] |
| 2 | gay sex aids men | ['gay', 'sex', 'aids', 'men', 'hiv', 'stigma', 'community', 'sexual', 'queer', 'about'] |
| 3 | children students schools school | ['children', 'students', 'schools', 'school', 'parents', 'child', 'kids', 'pediatric', 'campus', 'student'] |
| 4 | committee tedros international emergency | ['committee', 'tedros', 'international', 'emergency', 'global', 'pheic', 'countries', 'whos', 'ghebreyesus', 'adhanom'] |
| 5 | uk ukhsa nhs london | ['uk', 'ukhsa', 'nhs', 'london', 'security', 'agency', 'england', 'sexual', 'british', 'confirmed'] |
| 6 | minnesota case massachusetts patient | ['minnesota', 'case', 'massachusetts', 'patient', 'rare', 'contact', 'symptoms', 'virginia', 'department', 'officials'] |
| 7 | black white hispanic racial | ['black', 'white', 'hispanic', 'racial', 'data', 'new', 'us', 'disparities', 'mpox', 'vaccine'] |
| 8 | emergency declaration administration becerra | ['emergency', 'declaration', 'administration', 'becerra', 'biden', 'declare', 'response', 'declared', 'public', 'fentanyl'] |
| 9 | his my he her | ['his', 'my', 'he', 'her', 'basgoz', 'she', 'man', 'kwong', 'was', 'video'] |
| 10 | covid conspiracy pandemic theories | ['covid', 'conspiracy', 'pandemic', 'theories', 'covid19', 'our', 'fischman', 'you', 'car', 'we'] |

Upon interpretation, one could map some of the topics to the frames arising out of manual codebook development. In particular:

- Topic 2 → Sexual stigma and transmission routes
- Topic 7 → Racial disparities and stigmatising name
- Topics 4 and 8 → Epidemic preparedness and surveillance
- Topic 10 → Broader health issues

Note that if we were to utilize the topic model outcomes as a starting point for codebook development, we would still have to go through the clustered documents and develop a more exhaustive understanding of each frame, since the representative keywords for each topic only partly covers the breadth of the frame as understood via manual coding.

### A3.2.3 Generative large language models

LLM-based inductive thematic analysis

**Frame I: Vulnerable Populations**
- **Description:** Emphasizes the disproportionate impact of Mpox on specific groups, such as LGBTQ+ individuals and marginalized communities.
- **Problem definition:** The Mpox outbreak affects vulnerable populations disproportionately.
- **Causal interpretations:** Societal attitudes, lack of access to healthcare, and systemic inequalities are attributed to the disparities.
- **Moral evaluation:** Concern for stigmatizing vulnerable populations and a moral imperative to address health inequities.
- **Treatment recommendation:** Increased awareness, education, and targeted public health initiatives for marginalized communities.
- **Example:** "CDC data show that monkeypox cases are down dramatically in white men, but nearly 70% of cases are now being detected in Black or Latino men."

**Frame II: Risk Mitigation**
- **Description:** Focuses on individual behavior change and public education to prevent Mpox transmission.
- **Problem definition:** Individuals need guidance on minimizing risk of Mpox infection.
- **Causal interpretations:** Human behavior, social networks, and community factors are attributed to the outbreak.
- **Moral evaluation:** Concern for protecting individuals from harm and promoting personal responsibility.
- **Treatment recommendation:** Guidance and recommendations for individuals at risk, public education campaigns, and community engagement.

- **Example:** "People who have monkeypox should stay at home until the rash has healed…"

**Frame III: Public Health Emergency**
- **Description:** Frames Mpox as a serious public health concern requiring emergency response and funding.
- **Problem definition:** The rapid spread of Mpox cases globally causes concern for public health.
- **Causal interpretations:** Multiple factors, including viral transmission dynamics and human behavior, are attributed to the outbreak.
- **Moral evaluation:** Urgency and importance conveyed, with a focus on addressing the crisis.
- **Treatment recommendation:** Vaccination and prevention through behavioral changes are recommended.
- **Example:** "The World Health Organization declared Mpox a global health emergency due to its rapid spread."

**Frame IV: Stigma**
- **Description:** Highlights the fear of retribution and stigma against LGBTQ+ individuals due to Mpox.
- **Problem definition:** Fear of retribution and stigma against LGBTQ+ individuals due to Mpox.
- **Causal interpretations:** Societal attitudes and lack of understanding are attributed to the stigma.
- **Moral evaluation:** Moral imperative to address stigma and promote inclusivity.
- **Treatment recommendation:** Increased education, awareness, and support for marginalized communities.
- **Example:** "Noyes fears that because the virus has been linked to men who have sex with men, it could spark retribution against people who identify as LGBTQ+ in Humboldt County."

**Frame V: Vaccine Hesitancy**
- **Description:** Focuses on addressing vaccine hesitancy and promoting trust in public health officials.
- **Problem definition:** Limited vaccine uptake among high-risk groups and communities.
- **Causal interpretations:** Misinformation and mistrust of vaccines are attributed to the hesitancy.
- **Moral evaluation:** Importance of addressing misinformation and promoting vaccine confidence.
- **Treatment recommendation:** Efforts to combat misinformation, increase vaccine access, and promote trust in public health officials.
- **Example:** "Caulfield said the study is a reminder that public health officials need to adopt a range of responses to address misinformation on social media…"

**Frame VI: Resource Deficit**
- **Description:** Highlights the limited resources and infrastructure hindering public health response in rural areas.
- **Problem definition:** Limited resources and infrastructure hinder public health response in rural areas.
- **Causal interpretations:** Systemic issues and lack of funding are attributed to the resource deficits.
- **Moral evaluation:** Moral imperative to address resource inequalities and ensure equitable access to healthcare.
- **Treatment recommendation:** Increased funding, support for rural healthcare infrastructure, and targeted public health initiatives.
- **Example:** "The community health nurse responsible for distributing the vaccine from a state-run clinic in Winnemucca retired months ago, leaving a gap that was only filled with the arrival of a replacement nurse in October."

**Frame VII: Crisis Response**
- **Description:** Frames Mpox as a crisis requiring immediate attention and response.
- **Problem definition:** Mpox outbreak requires urgent response and action.
- **Causal interpretations:** Multiple factors, including viral transmission dynamics and human behavior, are attributed to the outbreak.
- **Moral evaluation:** Urgency and importance conveyed, with a focus on addressing the crisis.
- **Treatment recommendation:** Emergency funding, public health measures, and vaccination efforts are recommended.
- **Example:** "The U.S. is working to contain the largest monkeypox outbreak in the world…"

**Frame VIII: Global Health Cooperation**
- **Description:** Highlights the challenges in achieving global cooperation to address infectious disease threats.
- **Problem definition:** Challenges in achieving global cooperation to address Mpox outbreak.

- **Causal interpretations:** Systemic issues, lack of funding, and inadequate infrastructure are attributed to the challenges.
- **Moral evaluation:** Importance of addressing global health inequities and promoting international cooperation.
- **Treatment recommendation:** Increased funding, support for global health infrastructure, and targeted public health initiatives.
- **Example:** "…a small but noticeable proportion of people (2% to 3%) with monkeypox became very unwell…"

### Frame IX: Neurological Complications
- **Description:** Highlights the uncertainties about neurological complications associated with Mpox infection.
- **Problem definition:** Uncertainties about neurological complications associated with Mpox infection.
- **Causal interpretations:** Limited research data, complex interactions between viruses and host systems are attributed to the uncertainties.
- **Moral evaluation:** Concerns about the impact on public health and safety conveyed.
- **Treatment recommendation:** Continued research, monitoring, and transparent communication are recommended.
- **Example:** "…a small but noticeable proportion of people (2% to 3%) with monkeypox became very unwell…"

### Frame X: Government Intervention
- **Description:** Raises concerns about government involvement in personal lives and health decisions.
- **Problem definition:** Concerns about government involvement in personal lives and health decisions.
- **Causal interpretations:** Fear, mistrust, and perceived overreach of authority are attributed to the concerns.
- **Moral evaluation:** Importance of transparency, trustworthiness, and respect for individual autonomy conveyed.
- **Treatment recommendation:** Improved communication, public engagement, and collaborative governance recommended.
- **Example:** "'Any herb can clear that up' another man insisted. Wilson also faced a barrage of rejections…"

### Frame XI: Racial Disparities
- **Description:** Highlights the disproportionate impact of Mpox on men of color, particularly Black and Latino men.
- **Problem definition:** Disproportionate impact of Mpox on men of color, particularly Black and Latino men.
- **Causal interpretations:** Systemic racism and lack of access to healthcare are attributed to the disparities.
- **Moral evaluation:** Moral imperative to address racial health disparities conveyed.
- **Treatment recommendation:** Increased vaccination efforts and targeted public health initiatives for marginalized communities recommended.
- **Example:** "CDC data show that monkeypox cases are down dramatically in white men, but nearly 70% of cases are now being detected in Black or Latino men."

The LLM-based thematic analysis provides natural descriptions for each frame and a codebook, making it a superior approach over encoder-only transformer models. Upon interpretation, one could map some of the LLM frames to the frames arising out of manual codebook development. In particular:

- Vulnerable populations (I) and Stigma (IV) → Sexual stigma and transmission routes
- Racial disparities (XI) → Racial disparities and stigmatising name
- Global health co-operation (VIII) → Global relations
- Resource deficit (VIII) → Public health failure
- Public health emergency (III), Crisis response (VII) and Government intervention (X) → Epidemic preparedness and surveillance

It is promising that the LLM was able to output frames that are semantically similar to the gold standard. At face value, it may seem as if both the human coders and LLMs are interpreting the texts in a similar way, resulting nearly interchangeable codebooks. However, if the codebooks are truly interchangeable, the outcomes of frame detection using these two codebooks must be compatible. That is, if one were to use the same Llama model on the same test set with two different codebooks - one generated by humans and the other by Llama, with the Llama frames mapped onto semantically similar manual frames; they should yield the same results. However, as seen in the table below, the outcomes do not support the hypothesis that the codebooks are interchangeable. Put simply, even if the codebooks are semantically similar, their pragmatical interpretation may vastly differ.

Table A3: Frame detection metrics comparison for Llama with two different codebooks, one generated by Llama and the other by manual coders

| Frame | Accuracy | Kappa | F1 | Precision | Recall |
|---|---|---|---|---|---|
| Sexual stigma and transmission routes | 0.74 | 0.48 | 0.76 | 0.68 | 0.85 |
| Racial disparities and stigmatising name | 0.85 | 0.46 | 0.54 | 0.64 | 0.47 |
| Global relations | 0.73 | 0.33 | 0.45 | 0.3 | 0.92 |
| Public health failure | 0.82 | 0.52 | 0.62 | 0.48 | 0.88 |
| Epidemic preparedness and surveillance | 0.67 | 0.18 | 0.77 | 0.88 | 0.69 |
| Mean | 0.76 | 0.39 | 0.63 | 0.6 | 0.76 |

## A3.3. Frame detection

### A3.3.1 Human-generated labels

Table A4: Frame classification metrics for first round of human annotation.

| Frame | Accuracy | Kappa | F1 | Precision | Recall |
|---|---|---|---|---|---|
| Sexual stigma and transmission routes | 0.84 | 0.66 | 0.8 | 0.78 | 0.82 |
| Racial disparities and stigmatising name | 0.95 | 0.74 | 0.77 | 1 | 0.62 |
| Global relations | 0.86 | 0.39 | 0.47 | 0.38 | 0.6 |
| Public health failure | 0.88 | 0.64 | 0.71 | 0.75 | 0.68 |
| Epidemic preparedness and surveillance | 0.75 | 0.48 | 0.66 | 0.54 | 0.86 |
| Human interest stories | 0.95 | 0.64 | 0.67 | 1 | 0.5 |
| Broader health issues | 0.82 | 0.46 | 0.57 | 0.52 | 0.63 |
| Mean | 0.86 | 0.57 | 0.66 | 0.71 | 0.67 |

Table A5: Frame classification metrics for second round of human annotation.

| Frame | Accuracy | Kappa | F1 | Precision | Recall |
|---|---|---|---|---|---|
| Sexual stigma and transmission routes | 0.94 | 0.87 | 0.92 | 0.97 | 0.87 |
| Racial disparities and stigmatising name | 0.98 | 0.9 | 0.91 | 1 | 0.83 |
| Global relations | 0.95 | 0.71 | 0.74 | 0.78 | 0.7 |
| Public health failure | 0.96 | 0.86 | 0.88 | 1 | 0.79 |
| Epidemic preparedness and surveillance | 0.89 | 0.74 | 0.81 | 0.86 | 0.77 |
| Human interest stories | 0.99 | 0.94 | 0.94 | 1 | 0.89 |
| Broader health issues | 0.95 | 0.84 | 0.87 | 0.94 | 0.81 |
| Mean | 0.95 | 0.84 | 0.87 | 0.94 | 0.81 |

### A3.3.2 Discriminative language models

Table A6: Naïve Bayes: classification metrics with predictions evaluated against the labels produced by the first author after round 2 of manual annotation.

| Frame | Accuracy | Kappa | F1 | Precision | Recall |
|---|---|---|---|---|---|
| Sexual stigma and transmission routes | 0.79 | 0.55 | 0.72 | 0.73 | 0.71 |

| Frame | Accuracy | Kappa | F1 | Precision | Recall |
|---|---|---|---|---|---|
| Racial disparities and stigmatising name | 0.89 | 0.23 | 0.27 | 0.67 | 0.17 |
| Global relations | 0.89 | -0.02 | NaN | 0 | 0 |
| Public health failure | 0.85 | 0.4 | 0.48 | 0.7 | 0.37 |
| Epidemic preparedness and surveillance | 0.77 | 0.42 | 0.56 | 0.68 | 0.48 |
| Human interest stories | 0.9 | 0.24 | 0.28 | 0.4 | 0.22 |
| Broader health issues | 0.75 | 0.11 | 0.24 | 0.33 | 0.19 |
| Mean | 0.83 | 0.28 | 0.43 | 0.5 | 0.31 |

Table A7: BERT-base-cased: classification metrics with predictions evaluated against the labels produced by the first author after round 2 of manual annotation.

| Frame | Accuracy | Kappa | F1 | Precision | Recall |
|---|---|---|---|---|---|
| Sexual stigma and transmission routes | 0.79 | 0.57 | 0.74 | 0.7 | 0.79 |
| Racial disparities and stigmatising name | 0.95 | 0.77 | 0.8 | 0.77 | 0.83 |
| Global relations | 0.9 | 0 | NaN | NaN | 0 |
| Public health failure | 0.85 | 0.34 | 0.4 | 0.83 | 0.26 |
| Epidemic preparedness and surveillance | 0.75 | 0.41 | 0.59 | 0.6 | 0.58 |
| Human interest stories | 0.93 | 0.43 | 0.46 | 0.75 | 0.33 |
| Broader health issues | 0.75 | 0.33 | 0.49 | 0.43 | 0.57 |
| Mean | 0.85 | 0.41 | 0.58 | 0.68 | 0.48 |

Table A8: DeBERTa-v3-base: classification metrics with predictions evaluated against the labels produced by the first author after round 2 of manual annotation.

| Frame | Accuracy | Kappa | F1 | Precision | Recall |
|---|---|---|---|---|---|
| Sexual stigma and transmission routes | 0.82 | 0.61 | 0.75 | 0.79 | 0.71 |
| Racial disparities and stigmatising name | 0.9 | 0.45 | 0.5 | 0.62 | 0.42 |
| Global relations | 0.87 | 0.24 | 0.31 | 0.33 | 0.3 |
| Public health failure | 0.89 | 0.6 | 0.67 | 0.79 | 0.58 |
| Epidemic preparedness and surveillance | 0.8 | 0.52 | 0.65 | 0.7 | 0.61 |
| Human interest stories | 0.94 | 0.48 | 0.5 | 1 | 0.33 |
| Broader health issues | 0.79 | 0.22 | 0.32 | 0.5 | 0.24 |
| Mean | 0.86 | 0.45 | 0.53 | 0.68 | 0.46 |

### A3.3.3 Generative large language models

Table A9: Llama 3.1-8B-Instruct (8-bit quantized) zero-shot classification metrics with predictions evaluated against the labels produced by the first author after round 2 of manual annotation.

| Frame | Accuracy | Kappa | F1 | Precision | Recall |
|---|---|---|---|---|---|
| Sexual stigma and transmission routes | 0.7 | 0.43 | 0.7 | 0.56 | 0.92 |
| Racial disparities and stigmatising name | 0.94 | 0.75 | 0.79 | 0.69 | 0.92 |
| Global relations | 0.81 | 0.24 | 0.34 | 0.26 | 0.5 |
| Public health failure | 0.88 | 0.64 | 0.71 | 0.65 | 0.79 |
| Epidemic preparedness and surveillance | 0.68 | 0.36 | 0.61 | 0.49 | 0.81 |
| Human interest stories | 0.91 | 0.35 | 0.4 | 0.5 | 0.33 |
| Broader health issues | 0.76 | 0.45 | 0.6 | 0.46 | 0.86 |
| Mean | 0.81 | 0.46 | 0.59 | 0.52 | 0.73 |

Table A10: Llama 3.1-8B-Instruct (8-bit quantized) fine-tuned classification metrics with predictions evaluated against the labels produced by the first author after round 2 of manual annotation.

| Frame | Accuracy | Kappa | F1 | Precision | Recall |
|---|---|---|---|---|---|
| Sexual stigma and transmission routes | 0.69 | 0.42 | 0.7 | 0.55 | 0.95 |
| Racial disparities and stigmatising name | 0.93 | 0.72 | 0.76 | 0.65 | 0.92 |
| Global relations | 0.81 | 0.24 | 0.34 | 0.26 | 0.5 |
| Public health failure | 0.88 | 0.64 | 0.71 | 0.65 | 0.79 |
| Epidemic preparedness and surveillance | 0.75 | 0.47 | 0.67 | 0.57 | 0.81 |
| Human interest stories | 0.93 | 0.43 | 0.46 | 0.75 | 0.33 |
| Broader health issues | 0.76 | 0.45 | 0.6 | 0.46 | 0.86 |
| Mean | 0.82 | 0.48 | 0.61 | 0.56 | 0.74 |

Table A11: GPT-OSS-20B zero-shot classification metrics with predictions evaluated against the labels produced by the first author after round 2 of manual annotation.

| Frame | Accuracy | Kappa | F1 | Precision | Recall |
|---|---|---|---|---|---|
| Sexual stigma and transmission routes | 0.74 | 0.51 | 0.73 | 0.6 | 0.94 |
| Racial disparities and stigmatising name | 0.97 | 0.85 | 0.87 | 0.91 | 0.83 |
| Global relations | 0.73 | 0.25 | 0.35 | 0.22 | 0.88 |
| Public health failure | 0.87 | 0.66 | 0.74 | 0.59 | 1 |
| Epidemic preparedness and surveillance | 0.65 | 0.33 | 0.59 | 0.45 | 0.86 |
| Human interest stories | 0.95 | 0.75 | 0.78 | 0.64 | 1 |
| Broader health issues | 0.72 | 0.31 | 0.49 | 0.39 | 0.65 |
| Mean | 0.8 | 0.52 | 0.65 | 0.54 | 0.88 |

Table A12: Claude Sonnet 4 zero-shot classification metrics with predictions evaluated against the labels produced by the first author after round 2 of manual annotation.

| Frame | Accuracy | Kappa | F1 | Precision | Recall |
|---|---|---|---|---|---|
| Sexual stigma and transmission routes | 0.72 | 0.47 | 0.72 | 0.58 | 0.95 |
| Racial disparities and stigmatising name | 0.99 | 0.95 | 0.96 | 0.92 | 1 |
| Global relations | 0.92 | 0.51 | 0.55 | 0.62 | 0.5 |
| Public health failure | 0.89 | 0.68 | 0.75 | 0.67 | 0.84 |
| Epidemic preparedness and surveillance | 0.77 | 0.42 | 0.56 | 0.68 | 0.48 |
| Human interest stories | 0.98 | 0.88 | 0.89 | 0.89 | 0.89 |
| Broader health issues | 0.83 | 0.51 | 0.62 | 0.58 | 0.67 |
| Mean | 0.87 | 0.63 | 0.72 | 0.71 | 0.76 |

### A3.3.4 Error analysis and mitigation

Table A13: Llama 3.1-8B-Instruct (8-bit quantized) with LLM as judge: classification metrics with predictions evaluated against the labels produced by the first author after round 2 of manual annotation.

| Frame | Accuracy | Kappa | F1 | Precision | Recall |
|---|---|---|---|---|---|
| Sexual stigma and transmission routes | 0.65 | 0.34 | 0.66 | 0.52 | 0.89 |
| Racial disparities and stigmatising name | 0.97 | 0.87 | 0.89 | 0.8 | 1 |
| Global relations | 0.44 | 0.11 | 0.26 | 0.15 | 1 |
| Public health failure | 0.75 | 0.44 | 0.59 | 0.43 | 0.95 |
| Epidemic preparedness and surveillance | 0.34 | 0.03 | 0.48 | 0.32 | 1 |
| Human interest stories | 0.75 | 0.32 | 0.41 | 0.26 | 1 |
| Broader health issues | 0.31 | 0.05 | 0.36 | 0.22 | 0.95 |

| | Accuracy | Kappa | F1 | Precision | Recall |
|---|---|---|---|---|---|
| Mean | 0.6 | 0.31 | 0.52 | 0.39 | 0.97 |

Table A14: Llama 3.1-8B-Instruct (8-bit quantized) with human feedback: classification metrics with predictions evaluated against the labels produced by the first author after round 2 of manual annotation.

| Frame | Accuracy | Kappa | F1 | Precision | Recall |
|---|---|---|---|---|---|
| Sexual stigma and transmission routes | 0.69 | 0.42 | 0.7 | 0.56 | 0.92 |
| Racial disparities and stigmatising name | 0.96 | 0.81 | 0.83 | 0.83 | 0.83 |
| Global relations | 0.71 | 0.24 | 0.37 | 0.24 | 0.8 |
| Public health failure | 0.82 | 0.54 | 0.66 | 0.52 | 0.89 |
| Epidemic preparedness and surveillance | 0.58 | 0.25 | 0.57 | 0.42 | 0.87 |
| Human interest stories | 0.77 | 0.35 | 0.44 | 0.28 | 1 |
| Broader health issues | 0.72 | 0.38 | 0.55 | 0.42 | 0.81 |
| Mean | 0.75 | 0.43 | 0.59 | 0.47 | 0.87 |

Table A15: Llama 3.1-8B-Instruct (8-bit quantized) with decision trees: classification metrics with predictions evaluated against the labels produced by the first author after round 2 of manual annotation.

| Frame | Accuracy | Kappa | F1 | Precision | Recall |
|---|---|---|---|---|---|
| Sexual stigma and transmission routes | 0.5 | 0.15 | 0.6 | 0.43 | 0.97 |
| Racial disparities and stigmatising name | 0.95 | 0.75 | 0.78 | 0.82 | 0.75 |
| Global relations | 0.25 | 0.04 | 0.21 | 0.12 | 1 |
| Public health failure | 0.66 | 0.34 | 0.53 | 0.36 | 1 |
| Epidemic preparedness and surveillance | 0.36 | 0.05 | 0.5 | 0.33 | 1 |
| Human interest stories | 0.93 | 0.63 | 0.67 | 0.58 | 0.78 |
| Broader health issues | 0.72 | 0.37 | 0.54 | 0.41 | 0.81 |
| Mean | 0.62 | 0.33 | 0.55 | 0.44 | 0.9 |

Table A16: Llama 3.1-8B-Instruct (8-bit quantized) with confidence scores: classification metrics with predictions evaluated against the labels produced by the first author after round 2 of manual annotation.

| Frame | Accuracy | Kappa | F1 | Precision | Recall |
|---|---|---|---|---|---|
| Sexual stigma and transmission routes | 0.69 | 0.37 | 0.63 | 0.57 | 0.71 |
| Racial disparities and stigmatising name | 0.92 | 0.65 | 0.69 | 0.64 | 0.75 |
| Global relations | 0.79 | 0.22 | 0.32 | 0.24 | 0.5 |
| Public health failure | 0.9 | 0.7 | 0.76 | 0.7 | 0.84 |
| Epidemic preparedness and surveillance | 0.67 | 0.27 | 0.52 | 0.47 | 0.58 |
| Human interest stories | 0.91 | 0.35 | 0.4 | 0.5 | 0.33 |
| Broader health issues | 0.7 | 0.27 | 0.46 | 0.37 | 0.62 |
| Mean | 0.8 | 0.4 | 0.54 | 0.5 | 0.62 |

Table A17: Naive Bayes with Class Imbalance management: classification metrics with predictions evaluated against the labels produced by the first author after round 2 of manual annotation.

| Frame | Accuracy | Kappa | F1 | Precision | Recall |
|---|---|---|---|---|---|
| Sexual stigma and transmission routes | 0.72 | 0.43 | 0.67 | 0.6 | 0.76 |
| Racial disparities and stigmatising name | 0.9 | 0.49 | 0.55 | 0.6 | 0.5 |
| Global relations | 0.89 | 0.11 | 0.15 | 0.33 | 0.1 |
| Public health failure | 0.86 | 0.4 | 0.47 | 0.86 | 0.32 |
| Epidemic preparedness and surveillance | 0.76 | 0.48 | 0.67 | 0.59 | 0.77 |
| Human interest stories | 0.9 | 0.49 | 0.55 | 0.46 | 0.67 |
| Broader health issues | 0.78 | 0.26 | 0.39 | 0.47 | 0.33 |

| | Mean | 0.83 | 0.38 | 0.49 | 0.56 | 0.49 |

Table A18: BERT with Class Imbalance management: classification metrics with predictions evaluated against the labels produced by the first author after round 2 of manual annotation.

| Frame | Accuracy | Kappa | F1 | Precision | Recall |
|---|---|---|---|---|---|
| Sexual stigma and transmission routes | 0.75 | 0.45 | 0.64 | 0.71 | 0.58 |
| Racial disparities and stigmatising name | 0.84 | 0.03 | 0.11 | 0.17 | 0.08 |
| Global relations | 0.83 | 0.1 | 0.19 | 0.18 | 0.2 |
| Public health failure | 0.81 | 0.21 | 0.3 | 0.5 | 0.21 |
| Epidemic preparedness and surveillance | 0.71 | 0.34 | 0.55 | 0.53 | 0.58 |
| Human interest stories | 0.87 | 0.37 | 0.44 | 0.36 | 0.56 |
| Broader health issues | 0.78 | 0.08 | 0.16 | 0.4 | 0.1 |
| Mean | 0.8 | 0.23 | 0.34 | 0.41 | 0.33 |

Table A19: DeBERTa with Class Imbalance management: classification metrics with predictions evaluated against the labels produced by the first author after round 2 of manual annotation.

| Frame | Accuracy | Kappa | F1 | Precision | Recall |
|---|---|---|---|---|---|
| Sexual stigma and transmission routes | 0.72 | 0.45 | 0.69 | 0.59 | 0.84 |
| Racial disparities and stigmatising name | 0.52 | 0.04 | 0.23 | 0.14 | 0.58 |
| Global relations | 0.82 | 0.15 | 0.25 | 0.21 | 0.3 |
| Public health failure | 0.79 | 0.42 | 0.55 | 0.46 | 0.68 |
| Epidemic preparedness and surveillance | 0.73 | 0.43 | 0.63 | 0.55 | 0.74 |
| Human interest stories | 0.92 | 0.39 | 0.43 | 0.6 | 0.33 |
| Broader health issues | 0.75 | 0.35 | 0.51 | 0.43 | 0.62 |
| Mean | 0.75 | 0.32 | 0.47 | 0.43 | 0.58 |